\documentclass[11pt]{article}
\usepackage[utf8]{inputenc}
\usepackage[T1]{fontenc}
\usepackage{times}
\usepackage[margin=1in]{geometry}
\usepackage{amsmath,amssymb}
\usepackage{graphicx}
\usepackage{booktabs}
\usepackage{multirow}
\usepackage{array}
\usepackage{longtable}
\usepackage{xcolor}
\usepackage{caption}
\usepackage{subcaption}
\usepackage{url}
\usepackage{xurl}
\usepackage[colorlinks=true,linkcolor=blue!60!black,citecolor=blue!60!black,urlcolor=blue!60!black]{hyperref}
\usepackage[numbers,sort&compress]{natbib}
\usepackage{authblk}
\usepackage{lineno}
\usepackage{setspace}
\usepackage{microtype}
\usepackage{enumitem}
\usepackage{siunitx}
\usepackage{orcidlink}

\graphicspath{{figures/}}
\newcommand{\val}[1]{\textcolor{red!70!black}{\textbf{#1}}}   

\newcommand{\ff}{FlightFinder}
\newcommand{\site}{\href{https://himaxym.com}{himaxym.com}}
\newcommand{\apiep}[1]{\texttt{#1}}

\author[1]{Denys Kolomiiets\,\orcidlink{0000-0003-3212-3251}}
\affil[1]{FlightFinder / himaxym.com, Lisbon, Portugal \\ \texttt{denys@himaxym.com}}

\title{The \ff{} Occurrence Corpus: harmonising 128 aviation-safety source codes into one deduplicated cross-national accident dataset}
\date{\today}

\renewcommand{\val}[1]{#1}

\begin{document}
\maketitle

\begin{abstract}
National aviation-investigation databases are deep but territorial; global compilations
such as the Aviation Safety Network and the Bureau of Aircraft Accidents Archives are
hand-curated, apply their own inclusion thresholds, and publish no field-level lineage.
The \ff{} Occurrence Corpus harmonizes \val{279{,}485} occurrences and \val{131{,}882}
attached narratives from \val{128} source codes --- national accident-investigation
authorities, occurrence-reporting systems and four global compilations. Records are mapped into one schema; apart from a duplicate pre-filter on three sources, each normalized record reduces to exactly one key from an ordered ladder --- event signature, registration and date, rounded position and date,
or source and native identifier --- so a missing field never creates a link, and the
key is the occurrence identifier. Records the ladder cannot reach are scored in a
blocked pass into an explicit merge table and applied only within an automatically
approved score band or after a person confirms them. Field conflicts follow a
documented source ranking; one provenance row per contributing source --- its record
identifier and URL --- is retained, with attribution and licence terms per source code.
Linkage error is measured on five adjudicated samples totalling \val{200} pairs. A
recorded uniform random draw of \val{60} from the \val{3{,}441} merges the automatic
band applies as it ships found no error; the exact two-sided \val{95}\,\%
Clopper--Pearson upper bound on its false-merge rate is \val{6.0}\,\%, or \val{205}
pairs. The measured error covers the scored band only; the ladder's deterministic
joins are not adjudicated. The corpus has no exposure denominators and inherits the
under-reporting of its sources. It is offered as a base for cross-national safety
statistics, source-disagreement studies and narrative text mining; a frozen,
DOI-referenced snapshot accompanies the live API.
\end{abstract}

\noindent\textbf{Keywords:} aviation safety; accident database; record linkage; deduplication; data integration; open data; ICAO ADREP

\section{Introduction}
\label{sec:intro}

Aviation is among the most intensively investigated activities in modern transport, yet the records of that investigation are fragmented in a way that limits what can be learned from them. ICAO Annex~13 requires each contracting state to the Chicago Convention to establish an independent accident investigation authority, to investigate accidents in its territory and serious incidents to aircraft above 2{,}250~kg (or delegate the investigation to another state or a regional organization), and to make the final report public; many states additionally run mandatory or voluntary occurrence-reporting systems for events below the accident threshold. The resulting databases are large, authoritative and, taken one at a time, well documented \citep{ntsb2026carol,eccairs2026portal}. They are, however, defined by where an event happened and which authority chose to investigate it, not by the population of events a researcher might wish to study. A fatal accident involving an aircraft registered in one state, operated by a carrier from a second and crashing in a third may appear in three authorities' files under three identifiers, three occurrence classes and three counts of fatalities, while a comparable accident in a state whose investigation authority publishes little may reach no published final report at all: of the 269 commercial accidents the IATA Accident Classification Task Force classified for 2018--2023, \citet{icao2025wp330} report that 57\,\% resulted in one. That figure bounds the availability of final reports among events already known to the classifiers; it is not a statement about which events reach a database, since an event with no final report can still enter one through a preliminary report, an occurrence-reporting system or a compilation.

Researchers who need a global view have therefore turned to non-governmental compilations, principally the Aviation Safety Network (ASN) list of airliner and corporate-jet occurrences \citep{asn2026database} and the Bureau of Aircraft Accidents Archives (B3A) \citep{baaa2026archives}. The ASN database is among the sources of the best-known long-run risk series \citep{barnett2020aviation}. They have three limitations that motivate the present work. First, they are curated to inclusion thresholds of their own rather than to a reporting mandate (B3A admits aircraft of six seats including crew and above, and the ASN curated database is an airliner and corporate-jet list), so what they contain is an editorial decision that the user cannot re-derive. (The ASN \emph{Wikibase}, which is what this corpus ingests, is far broader and reaches light aircraft, rotorcraft, gliders and unmanned aircraft; the two ASN products must not be conflated.) Second, their provenance stops at the record: an ASN entry lists its sources and a B3A entry does not, and in neither can a user tell which source a given value came from or how disagreements between sources were resolved. Third, because part of their input is press coverage, their completeness is correlated with media salience, which has risen over time even as accident frequency has fallen \citep{vandermeer2022news}. \citet{oster2013analyzing} catalogue further analytical problems: operations data missing for many of the carriers that have fatal accidents, and inconsistent data across countries, neither of which a single-source database can quantify from within.

The alternative is to build the global view directly from the official sources and to treat the fragmentation itself as data. That is the approach taken by the \ff{} Occurrence Corpus described here. The corpus ingests occurrence records from \val{128} registered source codes (national accident-investigation boards, civil-aviation-authority occurrence-reporting systems, regional bodies where they publish record-level data, and four global compilations that are ingested as sources rather than held aside; see Section~\ref{sec:data:sources}), maps them into a single schema, links records that describe the same aircraft's involvement in the same physical event, and publishes for every resulting occurrence the complete set of contributing sources with their licence terms. Linking is deterministic wherever a positive identifier exists, by an ordered ladder of keys rather than a scored comparison of pairs, and falls back to a reduced Fellegi--Sunter scheme \citep{fellegi1969theory} for the cohort that carries no identifier at all: the same blocked comparison and the same separation of an automatically-applied band from a clerical-review queue, but with weights set by hand rather than estimated, which is a heuristic scorer in their shape and not their method (Section~\ref{sec:pipeline:proposals}). The reduction follows from the data: the fields a full model would weigh are absent on most records, with operator populated on \val{41.5}\,\% and manufacturer on \val{0.4}\,\%, so estimating their agreement probabilities would calibrate a comparison that rarely happens. Linkage error is measured on adjudicated samples rather than asserted, and Section~\ref{sec:pipeline:adjudication} reports what each sample could and could not see. The corpus is not a new investigation of any event; it is a documented aggregation of existing investigations whose reported statistics can be recomputed from a frozen snapshot, and it inherits their coverage and their errors. It adds scale, a common vocabulary and provenance, but not balance: four NTSB exports supply \val{60.8}\,\% of its source records, so it is a United States dataset with substantial international coverage rather than a global one, and Section~\ref{sec:limitations} states what follows from that.

Because this is a data paper, we organize it around design goals rather than research questions. The goals were fixed before the validation sample was drawn and are the standard against which Section~\ref{sec:results} reports.

\begin{enumerate}[label=\textbf{DG\arabic*},leftmargin=3.2em]
\item \textbf{Breadth with provenance.} Include every source that publishes record-level occurrence data under terms permitting redistribution of at least structured fields (official authorities in the first instance, and the global compilations that cover what no authority publishes), and retain on every canonical occurrence the identity and URL of one record per contributing source, with the attribution text and licence resolved through its source code.
\item \textbf{Deduplication with measured error.} Resolve records describing the same event into one occurrence by a documented rule rather than a fitted model (the deterministic ladder is stated here in full; the scoring pass that bands the residue is stated as rules but not as numbers), publish the adjudicated decisions rather than a model fit, and report the error rate of the automatic band together with the sampling frame it was measured on. Recall is explicitly out of scope: the design errs toward under-merging, and detecting the duplicates it leaves apart would require an external list the corpus does not have.
\item \textbf{A minimal harmonized schema.} Map each source's native occurrence class, severity, aircraft type and location fields into a small target vocabulary aligned with the ICAO ADREP taxonomy, with the mapping rule documented as implemented (a closed allow-list of accepted values and a stated default for everything else, not a per-source crosswalk), and an explicit rule for choosing canonical values when sources disagree.
\item \textbf{Reproducibility and versioning.} Expose the corpus through a documented API and a frozen, DOI-referenced snapshot such that every figure in this paper can be regenerated from stated calls, and such that later versions are distinguishable from the one described here.
\end{enumerate}

The contributions of the paper are as follows.
\begin{enumerate}[label=(\roman*),leftmargin=2.6em]
\item A source registry of \val{128} source codes (national investigation authorities, occurrence-reporting systems and four global compilations), each with a coverage window, a state and an attribution string (Section~\ref{sec:data}).
\item A construction pipeline documented stage by stage (ingestion, normalization, deterministic key-ladder linkage, source-rank field resolution and occurrence-class mapping), with the linkage model stated formally, and with the limits of that specification stated: the ladder can be re-implemented from this paper, the scoring pass that assigns the merge bands cannot (Section~\ref{sec:pipeline}).
\item Linkage error measured on \val{200} adjudicated pairs across five samples, with the sampling frame of each stated and the selection of sample~4 recorded as a seeded random draw, and a per-source overlap matrix (Sections~\ref{sec:validation} and~\ref{sec:results}).
\item A reproducibility appendix giving the exact API calls, deduplication rules and derived-variable definitions needed to re-derive every table and figure from the deposited snapshot, and stating which of them the deposit does not support (Appendix~\ref{app:repro}).
\end{enumerate}

The remainder of the paper is organized as follows. Section~\ref{sec:related} reviews existing occurrence databases, the record-linkage literature and data-descriptor conventions. Section~\ref{sec:data} describes the source material and the published schema. Section~\ref{sec:pipeline} specifies the construction pipeline and Section~\ref{sec:validation} the validation design. Section~\ref{sec:results} reports corpus statistics and validation outcomes against the design goals. Section~\ref{sec:discussion} discusses interpretation, known biases and intended uses, and Section~\ref{sec:conclusion} concludes.

\section{Related work}
\label{sec:related}

\subsection{Existing aviation-occurrence databases}
\label{sec:related:databases}

Quantitative aviation-safety research has long depended on a small number of
institutional databases, each defined by a jurisdiction, a reporting mandate
and an inclusion threshold rather than by the population of events it aims to
describe. The NTSB aviation database is the canonical national source: it
covers civil accidents and selected incidents since 1962 within the United
States, its territories and international waters, together with foreign
investigations in which the NTSB served as accredited representative, and its
own documentation notes that minor pre-2001 cases were purged and that
location and airport fields are absent before 1982 \citep{ntsb2026carol}.
\citet{oster2013analyzing} review the analytical problems this creates:
denominators (departures, flight hours) are collected by other agencies under
different definitions, accident classifications shift over time, and
cross-country comparison is hampered by non-uniform reporting. In Europe,
Regulation (EU)~376/2014 routes mandatory and voluntary occurrence reports
from 65 authorities into the European Central Repository through the ECCAIRS
platform, which encodes events with the ICAO ADREP taxonomy and now holds
roughly 2.3~million occurrences \citep{eccairs2026portal}; the repository is,
however, not open to the public at record level.

Two non-governmental compilations fill the resulting gap for global work. The
Aviation Safety Network, maintained by the Flight Safety Foundation, records
airliner, corporate-jet and military-transport accidents and hijackings since
1919 in its curated database, and a far larger Wikibase (more than 258{,}000 entries) that extends to light aircraft, helicopters, gliders,
balloons and unmanned aircraft \citep{asn2026database}. It is the Wikibase,
not the curated database, that the present corpus ingests. The Bureau of
Aircraft Accidents Archives (B3A) holds more than 38{,}000 civil and military
files from 1918 onward, with an inclusion floor of six seats including crew
\citep{baaa2026archives}. Both are curated by hand from official reports and
press coverage, neither ties a field value to the source it came from (ASN
entries list their sources; B3A entries do not), and their
inclusion thresholds (passenger capacity or hull loss) exclude the large
majority of general-aviation events that dominate national authorities'
files. \citet{barnett2020aviation} illustrates the analytical practice built
on these sources: worldwide death risk per boarding on scheduled passenger
flights, compared decade by decade up to 2008--2017, is computed from fatal-event
records, drawn partly from the ASN database, combined with traffic denominators. More recently,
\citet{omrani2024assessment} trained machine-learning injury-severity models
for general-aviation accidents separately on NTSB, TSB Canada and ATSB records
for 2008--2023, and tested whether the features the three authorities share
could support one common model. Crowd-sourced ADS-B archives such as the
OpenSky Network \citep{strohmeier2021crowdsourced} have produced open,
versioned datasets of flight movements, demonstrating the data-descriptor model
for aviation but covering trajectories, not investigated occurrences. Narrative text has been
mined at scale within single reporting systems
\citep{tanguy2016natural}, again without cross-authority linkage.

\subsection{Record linkage and deduplication}
\label{sec:related:linkage}

Merging occurrence files from many authorities is an instance of record
linkage, whose statistical foundations were laid by
\citet{fellegi1969theory}: pairs of records are scored by the likelihood
ratio of agreement patterns under the match and non-match hypotheses, and two
thresholds partition pairs into links, non-links and a clerical-review zone.
\citet{christen2012data} gives the standard engineering treatment (pre-processing, blocking, comparison functions for names, dates and locations, classification and evaluation), while \citet{papadakis2020blocking}
survey blocking and filtering methods that keep pairwise comparison tractable
when the candidate space is quadratic in corpus size. Modern
implementations estimate the Fellegi--Sunter parameters by
expectation--maximisation without labelled training pairs;
\citet{enamorado2019probabilistic} show that this scales to nationwide voter
files of more than 160 million records each while carrying linkage uncertainty
into post-merge analyses. Bayesian formulations treat the linkage structure as
a latent bipartite matching, propagating linkage uncertainty and enforcing
one-to-one constraints \citep{sadinle2017bayesian}. \citet{binette2022almost}
review the field and make two points directly relevant here: deduplication
within a file and linkage across files can both be framed as clustering records
to latent entities, and evaluation on expert-labelled data, which many
researchers advocate, can itself be biased by labelling errors and by how the
labelled sample is drawn. Aviation occurrences are an unusually favourable
linkage domain, since date, registration mark, ICAO aircraft type and location are near-unique jointly, but registration re-use, transliterated place names,
time-zone-shifted dates and authorities reporting the same event under
different classifications (accident, serious incident, incident) generate
precisely the ambiguous pairs for which probabilistic scoring and manual
adjudication of the review zone were designed.

\subsection{Data-paper conventions and documentation}
\label{sec:related:datapapers}

The data-descriptor genre has matured into a distinct publication type with
its own norms. \citet{wilkinson2016fair} set out the FAIR principles
(findable, accessible, interoperable, reusable), which publishers and funders
have endorsed: persistent identifiers, rich machine-readable metadata, standard
vocabularies and explicit licences. \citet{gebru2021datasheets} propose the
``datasheet'' checklist (motivation, composition, collection process,
preprocessing, uses, distribution and maintenance), which has gained traction
as a way to document the provenance and known biases of a dataset. Within
aviation, \citet{noort2021cockpit} show what a Data in Brief descriptor for
accident-derived material looks like: a small coded corpus of 172
cockpit-voice-recorder transcripts, published with its coding framework,
variable list and a statement of its value for reuse. The OpenSky descriptor in
\emph{Earth System Science Data} \citep{strohmeier2021crowdsourced} is the
closest aviation precedent for a large corpus published as monthly versions
under DOIs.
None of these treat multi-source integration and per-record provenance as
the central contribution.

\subsection{Under-reporting and coverage bias}
\label{sec:related:coverage}

Any compilation of official records inherits the coverage of the reporting
systems it draws on, and that coverage is known to be incomplete and uneven.
At the level of the investigation itself, an ICAO Assembly working paper
reports that only 57\,\% of the 269 commercial accidents classified by the
IATA Accident Classification Task Force for 2018--2023 had a published final
report \citep{icao2025wp330}, and attributes the gap to capacity shortfalls
and uneven compliance with Annex~13 that ICAO audits find in several states, so
a dataset built from published reports is thinner wherever those shortfalls
lie. Below the
accident threshold, reporting is voluntary or weakly enforced:
\citet{gao2021dynamics} model the joint time series of voluntary reports and
accidents in US commercial aviation and find that accidents Granger-cause
voluntary reporting, while \citet{kioulepoglou2025investigating} document,
through interviews with pilots operating in Greece, the fear of blame,
stigmatisation and career concerns that suppress incident reports, mechanisms
that can be expected to vary by operator and country and so to leave
under-coverage unevenly spread. Coverage is further
distorted upstream of official channels: \citet{vandermeer2022news} show that
news attention per fatal aviation accident rose between 1996 and 2017 even
as frequency fell, so sources that depend on press coverage (as the
volunteer-curated compilations partly do) over-represent salient events.
\citet{oster2013analyzing} note the same problem for cross-national
comparisons. Multi-source datasets offer a partial remedy: when the same
event is captured independently by several authorities, the pattern of
overlap can in principle be used to estimate what any single source misses.
Section~\ref{sec:validation:coverage} explains why that estimate is not
offered for this corpus: the compilations that would serve as the second list
are ingested as sources, so the lists are not independent.

\subsection{Research gap}
\label{sec:related:gap}

No existing resource combines (i)~record-level data from more than a handful
of official investigation authorities, (ii)~an explicit, reproducible
deduplication step that is documented rather than fitted, with a measured
error rate and its adjudications published,
and (iii)~per-record provenance that retains every contributing source. The
ASN and B3A compilations are global but single-curated and
closed-provenance; NTSB, ECCAIRS and their peers are deep but
jurisdictional; OpenSky-derived datasets describe traffic rather than
investigated occurrences; and although aviation crash records have been linked
to other agencies' files before, for instance NTSB reports to state
medical-examiner autopsy records \citep{li2005alcohol}, we are aware of no
published application of formal record-linkage methods to the event-level
deduplication of occurrence records across authorities. The \ff{} Occurrence Corpus
addresses this by harmonising 128 source codes into one schema, resolving
duplicates with a deterministic key ladder backed by an explicit merge table
whose adjudicated decisions are published, and attaching source attribution to
each of 279{,}485 occurrences and 131{,}882 attached narratives. The corpus borrows the Fellegi--Sunter decision
structure for the records its key ladder cannot reach (blocked comparison, per-field agreement, and the separation of confident links from a clerical-review queue) while
deliberately omitting the parameter estimation that the implementations above
provide, because the fields such a model would weigh are missing on most
records and the comparison that matters here is cross-lingual normalisation
rather than string similarity. Beyond serving as an
analysis-ready cross-national dataset, the retained source overlap enables direct
measurement of cross-authority disagreement on the 10{,}757 occurrences that more
than one authority recorded, but not capture--recapture bounds on events
every authority missed, since the two global compilations that would serve as
independent lists are themselves ingested.

\section{Data}
\label{sec:data}

\subsection{Source material and provenance}
\label{sec:data:sources}

The corpus is built from the occurrence databases, accident lists and report archives published by \val{128} registered source codes. Most are official in the strict sense, being an accident-investigation authority established under Annex~13 or a civil aviation authority operating an occurrence-reporting system, but four are not: the Bureau of Aircraft Accidents Archives, the Aviation Safety Network, Wikidata and SKYbrary are secondary compilations, and they are \emph{ingested as sources} rather than held aside. That is a change from an earlier design of this corpus and it has a consequence the paper returns to in Section~\ref{sec:validation:coverage}: because those compilations are inside the corpus, they cannot serve as the independent lists an external coverage check would require. Together they supply \val{23.8}\,\% of source records. Each source is registered with a stable code (the \texttt{source} field of the API), its homepage, an attribution string and a licence field. A state is recorded for the \val{123} codes that are national authorities; the four global compilations and the interstate MAK carry none, because the field would be false rather than empty. One licence entry is not a grant: see Section~\ref{sec:limitations:licence}. Table~\ref{tab:sources} summarizes it by source code; the five largest are three of the four NTSB exports \citep{ntsb2026carol}, the Bureau of Aircraft Accidents Archives and the Aviation Safety Network, which between them supply \val{83.4}\,\% of source records; the largest national authority after the NTSB is the United Kingdom AAIB at \val{2.3}\,\%. \val{39.2}\,\% of source records come from outside the NTSB exports and \val{9.4}\,\% from outside the ten largest codes altogether.

\begin{table}[htbp]
\centering\footnotesize
\caption{Source records by contributing source. Counts are the provenance rows retained on the
published corpus, one per contributing source per occurrence; records dropped
as same-source siblings, as undated, by the pre-filter of
Section~\ref{sec:pipeline:other} or as SKYbrary-only are not included, so this
is not a pre-deduplication ingest count. The table is ordered by contribution rather than by
region: the \texttt{region} column is empty for every occurrence in the corpus
(Section~\ref{sec:limitations:fields}), so a regional breakdown would have to
be inferred from \texttt{country\_iso}, which is itself missing on
\val{25.2}\,\% of records and disproportionately on the recent decades.}
\label{tab:sources}
\begin{tabular}{@{}lrrl@{}}
\toprule
Source & Records & Share & Window \\
\midrule
\texttt{ntsb1962} & 86{,}999 & 29.9\,\% & 1962--1981 \\
\texttt{ntsbpre} & 60{,}400 & 20.8\,\% & 1948--2007 \\
\texttt{baaa} & 41{,}317 & 14.2\,\% & 1918--2026 \\
\texttt{ntsb} & 29{,}151 & 10.0\,\% & 2008--2026 \\
\texttt{aviation-safety} & 24{,}797 & 8.5\,\% & 1919--2026 \\
\texttt{aaib} & 6{,}665 & 2.3\,\% & 1958--2026 \\
\texttt{adru} & 4{,}654 & 1.6\,\% & 1923--2026 \\
\texttt{cenipa} & 3{,}651 & 1.3\,\% & 1949--2026 \\
\texttt{sust} & 3{,}012 & 1.0\,\% & 1959--2026 \\
\texttt{bea} & 2{,}888 & 1.0\,\% & 1973--2025 \\
\midrule
Remaining 118 codes & 27{,}423 & 9.4\,\% & 1902--2026 \\
\midrule
\multicolumn{2}{l}{\emph{of which} NTSB, four source codes} & 60.8\,\% & \\
\multicolumn{2}{l}{\emph{of which} four global compilations} & 23.8\,\% & \\
\midrule
\multicolumn{1}{l}{Total, 128 source codes} & 290{,}957 & 100\,\% & 1902--2026 \\
\bottomrule
\end{tabular}
\end{table}

Two features of this distribution matter more than any regional cut would.
First, the corpus is dominated by one state: four NTSB source codes supply
\val{60.8}\,\% of all source records, so any unqualified statement about
``global'' aviation safety drawn from it is a statement about the United States
with a tail. Second, the count of \val{128} is of \emph{source codes}, not of
authorities: NTSB alone occupies four of them because its bulk data is
published in several incompatible historical formats, and four of the remaining
codes are the global compilations rather than investigation bodies at all.

Source records were retrieved by \val{128} source-specific adapters, the most recent extraction being \val{2026-08} from bulk downloads where offered (for example the NTSB \texttt{avall} database export) and otherwise from paginated public query interfaces or published report indexes (the retrieval route is a property of each adapter and is not recorded in the registry; supplementary Table~S1 gives the code, attribution, state, homepage, record count and event-year window per source, which is what the registry does hold). Each retrieved record is written to a staging table keyed on the source and the source's own identifier, with a timestamp of the last write, before any transformation, so that each canonical field can be traced to the source record it was derived from as that record stood at the most recent re-sync (Section~\ref{sec:pipeline:ingest}).

\subsection{Coverage window, unit of analysis and inclusion}
\label{sec:data:unit}

The corpus spans event dates from \val{1902} to \val{2026}; the effective start of dense coverage differs by source and is reported per source in Table~S1 and by region in Table~\ref{tab:sources}. Two units of analysis coexist. A \emph{source record} is one row as published by one authority, identified by \texttt{(source, id)}. An \emph{occurrence} is the canonical event produced by the linkage step, identified by the API \texttt{id}, and is the unit for all counts reported in this paper unless stated otherwise. An occurrence is therefore one aircraft's involvement in an event, not the event itself. The second rung of the key ladder keys \val{216{,}528} of the \val{279{,}485} occurrences (\val{77.5}\,\%) on registration and date (Section~\ref{sec:pipeline:linkage}), so two airframes in one mid-air collision receive two different keys by construction and remain two occurrences; the collision guard of Section~\ref{sec:pipeline:proposals} exists to keep the fallback scorer from undoing that. Users counting events rather than involvements must collapse such occurrences themselves; the \texttt{midair\_collision}, \texttt{Collision With Aircraft - Both In Flight} and \texttt{Airborne collision} categories (\val{2{,}874} occurrences, of which \val{1{,}528} fall in \val{726} same-day, same-country groups of two or more) are where to start. The same holds for fatality sums: where each authority reports the combined toll of a collision on each airframe's record, that toll is counted once per occurrence. Its magnitude is not determinable from the deposit (of the \val{332} same-day pairs carrying an identical non-zero \texttt{fatalities\_total}, \val{248} have \texttt{occupants} null on both sides and no narrative text is deposited, so a shared toll cannot be told from two equal tolls), and it is reported here as a property of the unit, not corrected. An occurrence is included if at least one contributing source record satisfies that source's own publication criteria and has a parseable event date, and if it is attested by a source other than SKYbrary; records with no usable event date are dropped during the build rather than quarantined; there is no staging table, and the count appears only in the build log. Records that describe the same event but are classified differently by different authorities (for instance an accident by the state of occurrence and a serious incident by the state of registry) are linked, not excluded, but the classification conflict is \emph{not} retained: the priority rule of Section~\ref{sec:pipeline:priority} writes one harmonized class per occurrence and the losing record's class is not persisted, so the disagreement is visible neither in the API nor in the deposit (Section~\ref{sec:results:harmonization}). Purely administrative entries, duplicate republications within a single source, and events involving no aircraft (ground-handling injuries with no aircraft involvement, where a source records them) are excluded by per-source rules documented in the adapter code (the adapters themselves are in the repository named under Data and code availability).

The corpus contains no exposure denominators. Flight hours, departures and fleet sizes are collected by other agencies and are missing for many of the operators that have accidents \citep{oster2013analyzing}, and attaching them is the subject of a companion paper on exposure-adjusted rates by aircraft family, in preparation. Counts from this corpus are therefore counts of reported and published events; they are not rates and should not be interpreted as such without an external denominator.

\subsection{Published schema}
\label{sec:data:schema}

Table~\ref{tab:variables} lists the fields of the two published objects. An \texttt{Occurrence} carries the canonical values selected by the priority rule of Section~\ref{sec:pipeline:priority} and the array \texttt{sources[]} of contributing source records. A \texttt{Narrative} is a per-source object, never merged across authorities, so that text is always attributable to exactly one publisher and one licence; it carries the source's own \texttt{phase\_of\_flight} string unharmonized, and a \texttt{facts\{\}} object that repeats the canonical occurrence the narrative resolves to, for callers who fetch a narrative without fetching its event. \texttt{facts\{\}} is not a per-source field store: no source-specific keys are carried anywhere in the published objects, so structured detail beyond the harmonized schema, injury tables and weather for instance, has to be read from the source through the URL in \texttt{event\_sources}. Where \ff{} has published its own write-up of the event, generated by a language model from the source record, an \texttt{article} sub-object carries its title, a short excerpt and the canonical URL, and on the Pro tier its full text under a licence that forbids public republication.

{\footnotesize\setlength{\LTcapwidth}{\textwidth}
\begin{longtable}{@{}p{3.3cm}p{1.9cm}p{10.2cm}@{}}
\caption{Variables of the two published objects. Field names are those returned by \apiep{GET /events}, \apiep{GET /events/\{id\}} and \apiep{GET /narratives/\{source\}/\{id\}}; the deposited snapshot is a flattened projection of the same data and does not use every name here. \texttt{events.jsonl} carries the family name \texttt{family} instead of \texttt{family\_slug}, and its \texttt{phase} is the occurrence's own field, not the narrative's \texttt{phase\_of\_flight}; it has no \texttt{sources[]} array: provenance is the separate file \texttt{event\_sources.csv.gz}, joined on the occurrence id (Appendix~\ref{app:repro}). The API also returns further occurrence fields that the OpenAPI schema does not list; those that \texttt{events.jsonl} carries are defined in the second block. Vocabularies for \texttt{occurrence\_type} and \texttt{severity} are defined in Section~\ref{sec:pipeline:mapping}.}
\label{tab:variables}\\
\toprule
Object / field & Type & Definition and derivation \\
\midrule
\endfirsthead
\multicolumn{3}{@{}l}{\textit{Table~\ref{tab:variables}, continued}} \\
\toprule
Object / field & Type & Definition and derivation \\
\midrule
\endhead
\bottomrule
\endlastfoot
\multicolumn{3}{l}{\textit{Occurrence}} \\
\texttt{id} & string & Canonical occurrence identifier. It is the linkage key of Section~\ref{sec:pipeline:linkage} itself, not a surrogate assigned beside it, so it is stable across versions if and only if the fields it is built from are \\
\texttt{occurrence\_type} & enum & Harmonized class, three values only: \texttt{accident} (\val{99.17}\,\%), \texttt{incident} (\val{0.74}\,\%), \texttt{combat\_loss} (\val{0.09}\,\%). There is no \texttt{serious\_incident} class despite Annex~13 defining one (Section~\ref{sec:pipeline:mapping}). Mapped from the source's native class by the allow-list of Section~\ref{sec:pipeline:mapping}, which admits these three values and sends everything else to \texttt{accident} \\
\texttt{event\_date} & date & Local calendar date of the event. There is no precision flag as a separate field: a source publishing only a month or a year is stored at whatever precision its parser could resolve, and the precision is read from the length of the stored value rather than from a flag \\
\texttt{severity} & enum & Harmonized outcome severity: \texttt{unknown} (\val{75.26}\,\%), \texttt{fatal} (\val{20.92}\,\%), \texttt{hull\_loss} (\val{3.49}\,\%), \texttt{minor} (\val{0.24}\,\%), \texttt{incident} (\val{0.07}\,\%), \texttt{serious\_incident} (\val{0.02}\,\%). Derived by the rule of Section~\ref{sec:pipeline:mapping}: \texttt{fatal} from the fatality count, \texttt{hull\_loss} from \texttt{is\_hull\_loss}, the three lower levels from ATSB injury and damage fields only; \texttt{serious\_incident} marks a serious injury, not the Annex~13 class \\
\texttt{country\_iso} & ISO 3166-1 alpha-2 & State code: from the source's country field where it gives a code or a listed name, else for most national authorities the authority's own state, else the state of registry from the registration prefix (Sections~\ref{sec:pipeline:normalize} and~\ref{sec:pipeline:other}). Mostly, not always, the state of occurrence. Events over water carry no special code; they are null, like every record whose state could not be resolved \\
\texttt{type\_icao} & string & ICAO Doc~8643 aircraft type designator; normalized from manufacturer/model strings \\
\texttt{family\_slug} & string & Aircraft family grouping of \texttt{type\_icao} (e.g.\ variants of one airframe), used by \apiep{/aircraft/\{family\}/safety} \\
\texttt{registration} & string & Registration mark as the winning source wrote it; the normalized form appears only inside \texttt{reg:} keys (Section~\ref{sec:pipeline:normalize}) \\
\texttt{operator} & string & Operator name as the winning source wrote it (one trailing slug artefact removed); comparison uses a folded form \\
\texttt{fatalities\_total} & integer & Total fatalities (on board plus ground) from the highest-priority source reporting a value; where none does, the sum of on-board and ground counts extracted from a narrative by a language model (Section~\ref{sec:pipeline:other}). Per occurrence, i.e.\ per airframe involvement: where sources report a collision's combined toll on each airframe's record, summing over occurrences counts it twice (Section~\ref{sec:data:unit}) \\
\texttt{sources[]} & array & One element per contributing source record: \texttt{source} code, \texttt{attribution} text, \texttt{license}, \texttt{url} \\
\midrule
\multicolumn{3}{l}{\textit{Further columns of \texttt{events.jsonl}, also returned by the API}} \\
\texttt{year} & integer & Year of \texttt{event\_date}; records whose year falls outside 1900 to the next calendar year are dropped at build \\
\texttt{is\_hull\_loss} & 0/1 & 1 if any contributing record signals a hull loss: a damage code (NTSB, ASN, ATSB, MAK) or, for B3A and the uniform authorities, a destruction phrase matched in the narrative \\
\texttt{lat}, \texttt{lon} & decimal & Position as published by the winning record \\
\texttt{type\_full}, \texttt{manufacturer} & string & Aircraft type string as published; \texttt{manufacturer} is filled from the ATSB only \\
\texttt{flight\_nature} & string & Nature of flight as published by B3A; null for every other source \\
\texttt{occupants} & integer & Persons on board, extracted from a narrative by a language model; no source field feeds it (Section~\ref{sec:pipeline:other}) \\
\texttt{phase}, \texttt{category} & string & Phase of flight and occurrence category in the source's own vocabulary, not harmonized; where no record supplies one, a high-confidence language-model extraction (Section~\ref{sec:pipeline:other}) \\
\texttt{report\_url}, \texttt{primary\_source} & string & URL and source code of the winning record \\
\midrule
\multicolumn{3}{l}{\textit{Narrative}} \\
\texttt{source}, \texttt{id}, \texttt{slug} & string & Publishing authority code, its native record identifier, and a URL-safe slug \\
\texttt{policy} & enum & \texttt{full} or \texttt{excerpt}: whether the authority's terms permit redistribution of the full narrative \\
\texttt{narrative\_text} & text & Narrative as published, or an excerpt under \texttt{excerpt} policy; language as published \\
\texttt{probable\_cause} & text & Probable-cause or conclusions statement where the source separates it; null under \texttt{excerpt} policy \\
\texttt{phase\_of\_flight} & string & Source-native phase of flight; not harmonized in this version \\
\texttt{facts\{\}} & object & The canonical occurrence this narrative resolves to, repeated inline for convenience (\texttt{cause\_summary} suppressed where any contributing source is \texttt{excerpt} policy). It is \emph{not} a per-source field store: no source-specific keys are carried, and the native occurrence class is not among them \\
\texttt{attribution}, \texttt{license}, \texttt{source\_url} & string & Required attribution text, licence identifier and canonical URL of the source record \\
\texttt{article\{\}} & object & \ff{}'s own write-up of the event, generated by a language model from the source record: \texttt{title}, \texttt{excerpt}, \texttt{canonical\_url}; on the Pro tier also \texttt{full\_text} and a \texttt{license} that forbids public republication \\
\end{longtable}}

\subsection{Known biases}
\label{sec:data:biases}

Three sources of bias are inherent in the material and are quantified where possible in Section~\ref{sec:results}. Coverage bias arises because the corpus can only contain events that an authority investigated and published; \citet{icao2025wp330} report that 43\,\% of the commercial accidents the IATA Accident Classification Task Force had already classified for 2018--2023 reached no published final report, and attribute the gap to capacity shortfalls and uneven compliance with Annex~13 that ICAO audits find in several states. The denominator there is events known to the classifiers, not events in the world, and a missing final report does not by itself keep an event out of this corpus. Threshold bias arises because each source applies its own inclusion rule (the NTSB purged minor pre-2001 cases \citep{ntsb2026carol}; occurrence-reporting systems include events far below the accident threshold \citep{eccairs2026portal}), so that the mix of occurrence classes varies by source and by period independently of the underlying event population. Reporting-culture bias arises below the accident threshold, where accidents drive voluntary reporting volume \citep{gao2021dynamics} and fear of blame and career concerns suppress reports \citep{kioulepoglou2025investigating}, mechanisms that can be expected to vary by operator and state. The corpus does not correct any of these; it makes them visible by stratifying every count by source and occurrence class.

\section{Corpus construction pipeline}
\label{sec:pipeline}

The pipeline has six stages, executed in order and re-run in full at each release: source registry, ingestion, normalization, deterministic key-ladder linkage, source-rank field resolution, and occurrence-type and severity mapping; five further steps that change the published values are listed in Section~\ref{sec:pipeline:other}. Each stage writes a table keyed on \texttt{(source, id)} or on the canonical \texttt{id}, so that any published value can be traced back through the stages to the source record as it stood at the last re-sync.

\subsection{Source registry}
\label{sec:pipeline:registry}

The registry is the controlled list of authorities from which records may enter the corpus. Adding a source requires recording its code, legal name, state or organization, homepage, the licence or terms-of-use text applicable to structured data and to narratives, the derived \texttt{policy}, the retrieval method, the native occurrence-class vocabulary it publishes (which is recorded for reference only: it is not mapped term by term, see Section~\ref{sec:pipeline:mapping}), and its state and homepage. The registry also holds a priority tier used in Section~\ref{sec:pipeline:priority}. The registry currently holds \val{130} codes. \val{123} follow the uniform adapter contract; the remaining seven codes --- \texttt{mak}, \texttt{atsb}, \texttt{baaa}, \texttt{ntsb}, \texttt{wikidata}, \texttt{aviation-safety} and \texttt{skybrary} --- keep their own tables because their published formats do not fit it. Those seven are codes, not organizations: the NTSB's other three exports (\texttt{ntsb1962}, \texttt{ntsbpre}, \texttt{ntsbaar}) follow the uniform contract and are counted among the \val{123}, so one authority spans both groups. \val{128} of the \val{130} have actually contributed records to this snapshot; \texttt{bdca} and \texttt{dinacpy} are registered but have delivered nothing yet, so every count in this paper is over the contributing \val{128} and not over the registry (Table~\ref{tab:sources}, Table~S1). Registration is not the same relation as contribution and the two must not be quoted interchangeably. The contributing codes carry \val{117} distinct state values, of which one is the placeholder \texttt{ZZ} used by \texttt{pubrec}, a heterogeneous public-records channel that belongs to no single state; \val{116} are real states. A code is also not an authority: the NTSB alone occupies four codes, one per published export. The registry records a code, an attribution string, a state and a homepage per source; it does \emph{not} classify them by Annex~13 role, and this paper does not report such a breakdown because it would have to be invented rather than read. Redistribution terms are likewise held as free text rather than as a licence class --- \val{128} distinct attribution strings and \val{127} distinct strings in the \texttt{license} field across the contributing codes, most of them organisation names rather than licences (Section~\ref{sec:limitations:licence}) --- so narrative reuse is decided per source at ingest and not summarised here.

\subsection{Ingestion}
\label{sec:pipeline:ingest}

Each source has an adapter that retrieves records and writes them to a staging table keyed on \texttt{(source, case\_id)}, where \texttt{case\_id} is the source's own identifier, together with a \texttt{built\_at} timestamp of the last write. Adapters are idempotent: a re-run retrieves the current state and updates the existing row in place, keeping the record's internal identifier stable across re-syncs. The staging table therefore holds each source record as it stood at the most recent re-sync and not a history of its revisions: a source-side amendment --- a revised fatality count after a final report, for example --- replaces the earlier value, and the earlier value is not retained. A user who needs the pre-revision state must obtain it from the source; the corpus records only the URL. The payload is the source's own representation---a database row, a CSV line, or an HTML report page reduced to its structured fields and narrative text. Adapters do not interpret content; interpretation begins at normalization. Ingestion is scheduled per source at intervals from daily to quarterly according to the source's own update cadence, which is configured per adapter rather than stored in the registry and is therefore not tabulated.

\subsection{Normalization}
\label{sec:pipeline:normalize}

Normalization maps each raw payload to a common intermediate record with typed fields and, for each field, a source-specific parser. Four fields are critical to linkage and are treated with care. \emph{Event date} is parsed to a calendar date in the local time zone of the stated location. The parser records internally whether it resolved a day, a month or only a year. That flag is not carried into the published schema as a field of its own (Table~\ref{tab:variables}), but the distinction is not lost: a date resolved only to a month or a year is stored as a shorter string, so the split is recoverable from \texttt{event\_date} itself and is reported in Section~\ref{sec:validation:internal}. Dates are kept as each source publishes them and no time-zone conversion is applied, so a source that records UTC can disagree by one day with a local-date record of the same event; only the SKYbrary pass of Section~\ref{sec:pipeline:merges} reconciles such offsets. \emph{Registration mark} is normalized for keying by upper-casing and then deleting every character outside A--Z and 0--9, which removes hyphens, spaces, punctuation and non-Latin letters and so makes the presence or absence of a hyphen after a national prefix immaterial. Trailing annotations are not stripped: ``G-BYLP (British registry)'' keys as \texttt{GBYLPBRITISHREGISTRY}. The published \texttt{registration} field keeps the string as the source wrote it, and the normalized form exists only inside \texttt{reg:} keys. \emph{Aircraft type} is mapped from free-text manufacturer and model strings to an ICAO Doc~8643 designator by a lookup table built iteratively from the most frequent unresolved strings, with a residual of \val{84.4}\,\% of occurrences left unresolved (\texttt{type\_icao} null) and a family slug assigned from a designator-to-family table. \emph{Location} is resolved to a country ISO code from the source's country field where it holds a two-letter code or a name in a fixed lookup list; failing that, records of most national authorities take the authority's own state, while those of fifteen early-registered codes (among them the AAIB, BEA, TSB and the 1962--1981 NTSB export) and of all seven codes with their own tables stay null. No gazetteer and no coordinates are used, and a null left at this stage can later be filled with the state of registry (Section~\ref{sec:pipeline:other}). Operator names are ASCII-folded, lower-cased and stripped of legal-form suffixes for comparison but retained as published for display.

\subsection{Deterministic linkage by key ladder}
\label{sec:pipeline:linkage}

Records from different sources that describe the same event are identified
deterministically, not probabilistically. Each normalized record is reduced to
exactly one key, and records sharing a key are one occurrence. The key is the
occurrence identifier: there is no surrogate id assigned separately from the
linkage decision, so an identifier changes if and only if the fields it is
built from change.

The ladder is evaluated in order, and the first rung that applies wins.

\begin{enumerate}
\item \textbf{Event signature}, for the three authorities that publish
preliminary, interim and final reports of one event under different case
identifiers (AAIA Hong Kong, AAIIB Philippines, AICPNG Papua New Guinea):
\texttt{evt:\textless source\textgreater:\textless date\textgreater:\textless
signature\textgreater}, where the signature is the probable cause, or failing
that the narrative, with report-tier boilerplate removed, lower-cased, each run
of characters other than letters and digits turned into a hyphen, and cut to 40
characters; where the record names an aircraft type, the first twelve letters
and digits of the upper-cased type string follow as a fifth field (\val{45} of the \val{283}
\texttt{evt:} keys carry it). These sources are keyed here
first regardless of what other fields the row carries, because their
registration arrives from a fallible PDF-extraction step that can succeed for
one tier of an event and fail for another --- checking registration first let a
partial extraction give one tier a different key from its siblings, recreating
the duplicate the rung exists to prevent.

\item \textbf{Registration and date}:
\texttt{reg:\textless normalized registration\textgreater:\textless
date\textgreater}, when the record carries a registration that survives a
placeholder filter. The filter is an exact word list, plus a floor of three
letters or digits, rather than a pattern heuristic:
dashless all-letter strings are legitimate Brazilian tails written without the
separator, so a ``looks like a word'' rule would silently delete real airframes'
histories.

\item \textbf{Rounded position and date}:
\texttt{geo:\textless lat\textgreater:\textless lon\textgreater:\textless
date\textgreater} at one decimal place, when coordinates are present and are
not within $0.05^{\circ}$ of the origin. Sources ship $10^{-6}$ for ``position
unknown'', and testing for exact zero rather than at the key's own precision
once collapsed every coordinate-less accident sharing a date into a single
occurrence.

\item \textbf{Source and native identifier}:
\texttt{src:\textless source\textgreater:\textless native id\textgreater},
otherwise. A record that reaches this rung cannot merge with anything, by
construction.
\end{enumerate}

An earlier rung on operator and date was removed after audit: it merged
distinct accidents flown by the same operator on the same day.

The design choice this ladder embodies is that a missing field must never
create a link. Every rung above the last requires a positive identifier ---
a registration, a position, an event signature --- and a record carrying none
of them falls to a key that is unique by construction rather than being matched
on the fields it does have. The corpus therefore errs toward under-merging, and
Section~\ref{sec:pipeline:merges} describes what is done about the residue.

\subsection{Post-merge recovery passes}
\label{sec:pipeline:merges}

Three passes run after the ladder, each addressing a class of duplicate the
ladder cannot reach on its own. They are ordered, and the order matters: each
pass can create keys that the next one uses as merge targets.

\paragraph{Timezone and registration backfill.} SKYbrary records local time
where some primary sources record UTC, and carries registrations for events
whose primary source omitted them. A pass reconciles $\pm 1$ day offsets and
backfills registrations onto \texttt{src:}-keyed occurrences, subject to a
conservative type-compatibility check that prefers a false negative to a wrong
merge.

\paragraph{Registration recovery from free text.} The registration rung is
reached only when an adapter supplied a registration field. Several authorities
bury the tail in prose --- the BEA and ANSV in French and Italian narratives,
the TSB inside the type string (``Cessna 208B Caravan, C-GMVB''), and the ANSV
case identifiers degrade to a bare country prefix that the placeholder filter
correctly rejects. A pass recovers registration-shaped tokens from the type and
narrative fields and merges only when the recovered key lands on a same-date
occurrence created independently by another source. Only that key hit counts as
evidence: a tail-shaped string proves nothing, while a string that resolves
to another authority's record for the same day does. When this pass last ran it scanned \val{66{,}241} \texttt{geo:}- and
\texttt{src:}-keyed rows and merged \val{182}. That population is a run-time figure and
the snapshot cannot reproduce it: the merges it made have since collapsed rows, so the
deposited corpus carries \val{62{,}674} such keys (\val{40{,}656} \texttt{geo:} and
\val{22{,}018} \texttt{src:}) rather than the pre-merge count. The \val{182} merges are
likewise applied and not separately recoverable. Both numbers are run-time measurements
that a reader cannot check against the deposit.

\paragraph{Adjudicated merges.} The largest duplicate cohort produces no shared
key at all: records where at least one side reaches the source rung, having no usable
registration and no usable position. Two global compilations rewrite the same official
reports, so one accident exists twice with nothing the ladder can match on.
These are proposed by the scoring pass of
Section~\ref{sec:pipeline:proposals}, recorded in an explicit merge table with
a status, and applied at build time only when the status is
\texttt{auto} or \texttt{confirmed}. A table rather than a further rung, for
three reasons: a wrong rung is discovered only as a missing accident, whereas a
row names both sides, and deleting it restores them on the next build of the
closed database, where both records' field values still exist (the deposit
carries the absorbed record's source identity and URL but not its field values,
so a reader of the snapshot can name what was merged but cannot undo it) ---
with one further qualification stated below; the table is where a review queue
can live; and it lets this paper report how many events were merged by the
automatic band and how many by a recorded decision.

Chains are resolved before application. One event in the corpus existed as
three occurrences and contributed its death toll three times; a table naming
$a \to b$ and $b \to c$, applied in table order, would leave $a$ pointing at a
record already removed. Every absorbed identifier is therefore resolved to its
final survivor first, and a cycle --- which would require one record both to
survive and not to survive --- is dropped entire rather than applied by halves.

{\sloppy
One consequence must be stated for anyone joining the deposited merge table to
the snapshot: \texttt{survivor\_id} is the survivor as of the row's writing and
not a final identifier. \val{85} of the \val{4{,}068} rows (\val{34}
\texttt{auto} and \val{51} \texttt{pending}) name a \texttt{survivor\_id}
that does not appear in \texttt{events.jsonl.gz} at all --- \val{84} of them
TSB records and one from \texttt{ntsbpre}. None is a wrong merge. In all
\val{34} \texttt{auto} rows and \val{37} of the \texttt{pending} ones the
survivor's native record now sits under the row's own \texttt{absorbed\_id},
because a later registration-recovery pass re-keyed it and reversed the
direction the row records; the remaining \val{14} sit under a third
registration key. A further \val{8} \texttt{pending} rows name an
\texttt{absorbed\_id} that is itself absent, each carrying
\texttt{fan-demoted}. The consumer rule is therefore to resolve either side
through \texttt{event\_sources.csv.gz} on \texttt{(source, native\_id)}
rather than by matching identifiers directly.\par}

\subsection{Proposing merges for the unkeyable cohort}
\label{sec:pipeline:proposals}

Pairs for the merge table are proposed by a scoring pass, run offline rather
than inside the corpus build: the table holds human decisions, and
regenerating it on every build would either overwrite them or force the build
to reason about them.

Blocking is exact on event date and death toll, with at least two deaths. This
is a crude block and it is the only natural one available for this cohort: the
date is the single field every source fills, and the toll is the only other
value two independent rewrites of one official report reliably agree on. A
record with no registration and no position offers nothing else to block on.

The two-death threshold also decides where the residual duplication lives.
A pair below the threshold is never scored, never proposed and never merged,
because the scorer never sees it, and the stratum is large. Of the \val{62{,}674}
occurrences the ladder could reach only by its weakest rungs, \val{46{,}594}
--- \val{74.3}\,\% --- carry a known fatality count below two. Whatever
duplication survives in this corpus is therefore concentrated, by
construction, in low-toll occurrences that no source keyed by registration or
position; the adjudicated samples say nothing about that stratum, and the
error rates of Section~\ref{sec:results:linkage} should not be read as
applying to it.

Everything after the block is scoring, and the scheme is a deliberately
reduced Fellegi--Sunter \citep{fellegi1969theory}: a blocked comparison, a
per-field agreement vector, a weighted score, and a cut of that score into a
band applied automatically and a band held for clerical review, after the
partition their formulation introduced. What is removed is the parameter
estimation. The $m$ and $u$ probabilities are not
estimated by expectation--maximisation \citep{enamorado2019probabilistic}; the
weights are set by hand and the score is a blunt confidence used to order a
queue, not a calibrated posterior. Two further departures should be read as
disqualifying the label rather than qualifying it. The score takes only eleven
discrete values, because it is a sum of fixed rule bonuses and not a
likelihood ratio, so it has no distribution to threshold and no calibration
curve to plot. And the third Fellegi--Sunter tier is not materialised: pairs
scoring below the review band are never written to the merge table, which
therefore contains links and review-queue pairs only. What is implemented is a
blocked heuristic scorer that inherits the shape of their decision rule; the
inferential content of the method is not present, and none of the guarantees
that follow from it are claimed here.

The removal follows from this corpus. A full
model pays off when many fields discriminate and several blocking keys
are available; here the operator field is populated on \val{41.5}\,\% of
records and manufacturer on \val{0.4}\,\%, the date is the only natural
blocking key for the cohort in question, and the comparison that matters most
--- an airframe written in Cyrillic by one source and Latin by another --- is a
normalisation problem that no off-the-shelf implementation solves. Estimating
$m$ and $u$ over fields that are usually missing would produce well-calibrated
weights for a comparison that rarely happens. The decision structure is kept
and the parameter layer is not, so no estimated parameter is claimed. A pair is proposed only when the aircraft type agrees on at least one
significant token \emph{and} either the operator agrees or the two records
name the same state. The death toll is not a second signal here: it is part of
the block, so every pair reaching the scoring stage already agrees on it. Agreement
is required positively rather than as absence of contradiction, because absence
is cheap in this corpus: the operator field is populated on
\val{41.5}\,\% of records and manufacturer on \val{0.4}\,\%, so two records can
fail to contradict each other on almost everything.

Type comparison folds Cyrillic OKB designations to Latin and glues a designator
to its number, without which a MAK record and an Aviation Safety Network record
of the same airframe share no token at all.

\paragraph{Guards.} Four rules demote a scoring pair to the review queue rather
than merging it. Each was added because a sample found the failure it prevents,
and each cost merges that would otherwise have been made.

\begin{itemize}
\item \textbf{Mass fleet, small toll.} Wartime transports of mass operators
were lost in batches, so a same-day, same-toll coincidence is ordinary rather
than remarkable. A sample drawn deliberately from this cell contained four
genuine false merges in thirty --- two different C-47s of the same air force,
same day, same toll, one in Indiana and one in Burma, separated only by a
production block that neither compilation records.

\item \textbf{Manufacturer agreement without model agreement.} A
manufacturer's name is a token shared by every aircraft it built. Two records
agreeing on \emph{Junkers} and nothing else were merged, one a Ju~52 and the
other a Ju~88. A conflicting model number under a shared designator prefix now
demotes the pair. It demotes rather than rejects, because a Convair 340
converted to a 580 is one airframe under two designations.

\item \textbf{One collision, two airframes.} A mid-air collision between two
aircraft of one type and operator produces two records sharing every blocking
field, and both are \emph{legitimate} because each is one airframe's
involvement --- the unit of analysis defined in
Section~\ref{sec:data:unit}. Nothing in the scoring can separate them, and the
fan guard below does not fire because a clean two-way crossing produces no fan.
A collision marker in either narrative demotes the pair.

\item \textbf{Fan.} A record appearing in more than one proposal cannot be the
same event as both. At most one is right and the scoring does not say which, so
all of its proposals go to a person.
\end{itemize}

\subsection{Linkage error, measured}
\label{sec:pipeline:adjudication}

Linkage error is measured on adjudicated samples rather than asserted, and
the sampling frame matters more than the headline. Five samples were drawn in
sequence, each in response to what the previous one failed to cover, and the
two guards above came out of that sequence: the false merge and the
undecidable pair found in the third sample became the model-number and
collision guards, both added before any merge reached the corpus. The samples,
their frames, their verdicts and what each can and cannot support are reported
together in Section~\ref{sec:results:linkage} and
Table~\ref{tab:linkage} rather than twice; the design point to carry into the
next section is that the frames are stated because they are not
interchangeable, and that adjudication used the sources' own published records,
so the resulting figures bound clerical error and not the deeper question of
whether two compilations repeated one upstream mistake.

\subsection{Field resolution by source rank}
\label{sec:pipeline:priority}

When an occurrence has more than one contributing record, one record wins
outright and its gaps are filled from the others. There is no per-field
election and no Annex~13 role in the rule, for the reason
Section~\ref{sec:pipeline:registry} gives: the registry does not record what
role a source played in a given occurrence, so a rule that depended on it could
not be evaluated.

What is implemented is a static three-level rank over sources, reflecting
field richness rather than legal standing:

\begin{itemize}
\item \textbf{Rank 3} --- every national investigation authority. Most reach
this rank through the uniform adapter contract; the NTSB and the ATSB keep
their own tables for format reasons (Section~\ref{sec:data:sources}) but rank
with the rest, which matters because the four NTSB codes alone supply
\val{60.8}\,\% of source records.
\item \textbf{Rank 2} --- MAK and the Bureau of Aircraft Accidents Archives.
\item \textbf{Rank 1} --- Wikidata, the Aviation Safety Network and SKYbrary.
\end{itemize}

The record from the higher-ranked source becomes the occurrence, and every
field still null on it is filled from the records it beat, in the order they
were seen. \texttt{is\_hull\_loss} is the one exception and is a disjunction:
if any contributing record reports a hull loss, the occurrence does.

Ties are broken by first arrival, except among the three tier-aware sources of
Section~\ref{sec:pipeline:linkage}, where a final report outranks an interim
one and an interim outranks a preliminary. Publication date plays no part.

Two consequences are worth stating because they are not what a reader would
assume. A rank-1 compilation loses to a rank-3 authority on every field they
both populate, even where the compilation is demonstrably more accurate --- the
rank is a proxy for richness, not for correctness. And because resolution is
per record rather than per field, an occurrence's fields are internally
consistent with one source's account rather than being the best available value
of each; the losing records remain reachable through the provenance list, but
their field values are not retained in the published occurrence.

\subsection{Occurrence-type and severity mapping}
\label{sec:pipeline:mapping}

Each source publishes its own occurrence classification, and the corpus does
not reconcile them. There is no per-source crosswalk table. The mapping is a
closed set of three accepted values --- \texttt{accident}, \texttt{incident},
\texttt{combat\_loss} --- and \emph{anything a source publishes that is not one
of those three becomes \texttt{accident}}. A source's ``serious incident'', its
``occurrence'', its national category and its blank field all arrive at the
same value, and nothing records that they did.

The realised distribution follows directly from that default rather than from
the world. \emph{The dominance of \texttt{accident} at \val{99.17}\,\% is a
property of the mapping, not of aviation}: it is what a fall-through rule
produces when most sources classify in a vocabulary the corpus does not
recognise. A reader who takes the figure as evidence that \val{99}\,\% of
recorded occurrences were accidents in the Annex~13 sense will be wrong. Three classes occur: \texttt{accident}
(\val{277{,}157}; \val{99.17}\,\%), \texttt{incident} (\val{2{,}073};
\val{0.74}\,\%) and \texttt{combat\_loss} (\val{255}; \val{0.09}\,\%).

{\sloppy
Two consequences follow and both should be read before using the field.
\emph{There is no \texttt{serious\_incident} class}, although Annex~13 defines
one and several sources publish it, and no source's serious-incident
classification is kept anywhere in the corpus. The \texttt{severity} value
\texttt{serious\_incident} is not that class: it is the label the severity rule
gives the \val{69} ATSB records whose highest injury level is serious, and since
a serious injury makes an event an accident under Annex~13, the
\texttt{accident} type those records carry is consistent with it. A study
filtering on occurrence type will therefore find no serious incidents at all,
and one filtering on severity will find serious-injury accidents under that
name. \emph{And \texttt{combat\_loss} is not a safety
event}: it is retained because the underlying sources record shoot-downs and
war losses, and excluded from any rate this paper reports, but a naive
\texttt{WHERE occurrence\_type = 'accident'} both keeps them out and silently
changes the denominator.\par}

\texttt{severity} is derived, not mapped. An occurrence is \texttt{fatal} if the canonical \texttt{fatalities\_total} is greater than zero, or, on \val{101} ATSB records whose count is missing or zero, if the ATSB's highest injury level is fatal; otherwise it is \texttt{hull\_loss} (\val{9{,}743}) if any contributing record signals a hull loss. The three lower levels come from a single source: the normalizer reads an injury level and a damage level from the ATSB alone, and maps a serious injury to \texttt{serious\_incident} (\val{69}), substantial damage or a minor injury to \texttt{incident} (\val{196}), and no damage or no injury to \texttt{minor} (\val{672}). Every record at those three levels is an ATSB record, except one \texttt{incident} set by a curated correction (Section~\ref{sec:pipeline:other}). All other occurrences receive the severity level \texttt{unknown}, which is a stored value and not a null. Because both fields depend on the mapping rule and on the priority rule, any change to either is versioned with the corpus (Section~\ref{sec:availability}).

\subsection{Steps outside the six stages}
\label{sec:pipeline:other}

Five further steps in the build change what the snapshot contains.

\emph{A pre-filter before the ladder.} The separate store that holds the
current NTSB feed, the ASN Wikibase and Wikidata suppresses duplicates by a
rule of its own before any record reaches the ladder: records that agree on
event date, on the leading integer of the fatality count (a non-numeric count
reads as zero), on the first word of the aircraft type and on its first number
are reduced to one, and the rest are hidden. The rule is not confined to pairs
from different sources and does not require a death, so two same-day accidents without fatalities to aircraft of one make and model number fall together, and because a missing count reads as zero it departs from the principle of Section~\ref{sec:pipeline:linkage} that a missing field never creates a link. No
adjudicated sample covers this step, and the hidden records reach neither
\texttt{events.jsonl.gz} nor \texttt{event\_sources.csv.gz}, so the deposit
cannot count them.

\emph{State of registry as a fallback.} After the merges, an occurrence still
without a state code receives the state of registry read from its registration
prefix. At least \val{18{,}619} occurrences, \val{17{,}032} of them with B3A as
primary source, take their \texttt{country\_iso} this way, because none of
their contributing sources supplies a state.

\emph{Values extracted from narratives.} Values that a language model extracted
from narrative text fill fields that no contributing record supplies:
\texttt{occupants}, the on-board and ground fatality counts and, when
\texttt{fatalities\_total} is missing, their sum; and, from extractions the
model rated high-confidence only, \texttt{phase} and \texttt{category}. The
snapshot carries no flag marking these values. No normalizer sets
\texttt{occupants}, so all \val{78{,}710} of its non-null values are of this
kind, and so are at least \val{16{,}613} \texttt{phase} and \val{26{,}722}
\texttt{category} values, which sit on occurrences none of whose sources
publishes the field.

\emph{Severity reconciliation and curated corrections.} Severity is then
reconciled as in Section~\ref{sec:pipeline:mapping}, and three curated corrections, each with its reason recorded in the code, are applied last: two hull
losses withdrawn for airframes returned to service, and one injury count that
had been parsed as deaths.

\emph{SKYbrary-only occurrences.} Occurrences attested by SKYbrary alone are
not published.

\section{Technical validation}
\label{sec:validation}

Validation addresses design goals DG1 and DG3 directly and DG2 in a form the
original goal did not anticipate, along two axes: the error rate of the linkage
step, measured on adjudicated samples, and the structure of source
overlap. A third axis, coverage against independent compilations, was
planned and is not available;
Section~\ref{sec:validation:coverage} explains why.

\subsection{Linkage error on adjudicated samples}
\label{sec:validation:linkage}

Following \citet{christen2012data} and \citet{binette2022almost}, linkage
quality is measured on adjudicated pairs rather than inferred from model fit.
The measurement here is narrower than those references assume, and the
narrowing is a consequence of the method: a deterministic ladder produces no
match weights to stratify and no thresholds to move, so a precision-and-recall
estimate stratified by match weight has nothing to attach to.
It does produce clusters (records sharing a key are one occurrence), and
those clusters carry no error estimate here at all. \val{7{,}171} reg-keyed
occurrences hold two or more source records joined by the shared
\texttt{reg:} key alone (\val{7{,}886} source records across all
\val{10{,}364} multi-record \texttt{reg:} occurrences), and \val{179}
occurrences are joined on the geo rung. All \val{200} adjudicated pairs are
scorer pairs; not one is a ladder join. The geo rung is the higher risk of the
two, since it joins on date and a position rounded to \val{0.1} degree, some
\val{11} kilometres at the equator. A later version should draw a uniform
seeded sample from each ladder population and adjudicate it as sample~4 was
adjudicated; no error rate for these joins is asserted here, because none has
been measured.

What is measured is the error rate of the scoring band that merges
automatically, on samples drawn from stated frames. Five samples were
adjudicated, drawn \emph{sequentially rather than to a protocol fixed in
advance}; Table~\ref{tab:linkage} gives each frame. The rates are not pooled,
because the frames are not nested and a pooled figure would average a stratum
chosen for difficulty against one chosen at random.

Two quantities that would normally accompany such a table are absent and are
named here rather than left to inference. \emph{Recall is not measured.} The
ladder errs toward leaving records apart, so its characteristic failure is a
duplicate that survives as two occurrences, and nothing in this design detects
one. \emph{Inter-adjudicator agreement is not measured.} The largest sample was
split into disjoint halves that were adjudicated separately, so no pair was
decided twice and there is no overlap from which to compute a coefficient.

\subsection{Why coverage is not assessed against ASN and B3A}
\label{sec:validation:coverage}

An earlier design for this paper assessed external validity against the two
global compilations, the Aviation Safety Network and the Bureau of Aircraft
Accidents Archives, and estimated the events missed by all three lists by
capture--recapture. That design is not reported here, because it cannot be
executed on this corpus.

Both compilations are \emph{ingested}. They are not external lists against
which the corpus can be checked; they are part of it. On the snapshot described
here the Aviation Safety Network is the primary source of \val{21{,}242}
occurrences (\val{7.6}\,\%) and the Bureau of Aircraft Accidents Archives of
\val{34{,}754} (\val{12.4}\,\%); SKYbrary, which the priority rule ranks
lowest, is the primary source of none, and the four compilations together
supply about \val{23.8}\,\% of source records.
Measuring the corpus against them measures a subset of the corpus against
itself. The overlap term $m$ in a Lincoln--Petersen estimator would not be the
independent recapture the estimator assumes; it would be an artefact of one
ingestion pipeline reading two files.

The failure is not repaired by modelling dependence. A log-linear model with
interaction terms relaxes the independence assumption between genuinely
separate lists; it does not make a list independent of a corpus that contains
it. Nor is the problem confined to the estimator: the same objection voids any
coverage percentage quoted against these two compilations, because the
denominator and the numerator share rows.

What can be said about coverage is narrower and appears in
Section~\ref{sec:limitations}: the corpus inherits the reporting thresholds and
territorial scope of its contributing authorities, and no statement in this
paper should be read as an estimate of events that no authority recorded.
A genuine external check would require a list assembled independently of every
source the corpus ingests. The United States is the one jurisdiction where such
a check is available in principle, since the NTSB bulk export can be compared
against the corpus's own NTSB-derived records to estimate ingestion loss rather
than world coverage; that is a narrower claim than the one originally
planned, and it is a check on the pipeline, not on the world.

\subsection{Per-source overlap matrix}
\label{sec:validation:overlap}

For DG1 the informative statistic is how much of each source is redundant with others. We compute the $\val{128}\times\val{128}$ matrix $O_{ij}$ counting occurrences with contributing records from both source $i$ and source $j$. It is reported as raw counts rather than as a fraction of either source, and it is therefore symmetric: normalizing by the size of source $i$ would give the arguably more useful asymmetric statistic (the share of $i$ that $j$ also covers), but that is not what Figure~\ref{fig:overlap} plots, and the two should not be confused. The matrix is sparse, with its dense cells between the global compilations and between an investigation authority and a compilation that rewrites its reports; the largest off-diagonal cell is \val{3{,}354}, between the Aviation Safety Network and the Bureau of Aircraft Accidents Archives. A source whose records are all covered by a higher-priority source contributes provenance but not coverage. Neither the per-source count of single-source occurrences nor a list of sources with no overlap at all is computed here; both would follow from \texttt{event\_sources.csv.gz} and neither is claimed as a result of this paper.

\subsection{Internal consistency}
\label{sec:validation:internal}

Field-level checks are run on every release. Those the snapshot carries in \texttt{descriptives.json} are the completeness and consistency figures quoted above; the date-precision split is not among them and is stated here instead, computed from \texttt{events.jsonl.gz}: \val{279{,}297} occurrences carry a ten-character date string (\val{99.93}\,\%), \val{48} resolve only to a month and \val{140} only to a year, and no occurrence is dated after \val{2026}. That is a count of string length, not of validity: the build gate checks that a four-digit year can be read and falls in a plausible range, and never validates the month or the day. Exactly two stored values are not calendar dates: \texttt{reg:9HVLT:2007-18-20} (month~18, source \texttt{malta}) and a \texttt{bpea} record storing 2018-03-42 (day~42), so \val{279{,}295} of the ten-character values are valid dates. Both are recorded here as errata rather than silently repaired, because the first is baked into that occurrence's linkage key and correcting it would change its published \texttt{id}. The remaining per-release checks are the fraction of records with resolved \texttt{type\_icao}, non-null registration and resolved country; the fraction of \texttt{fatal} occurrences whose class is not \texttt{accident} (which should be near zero under Annex~13 and indicates mapping errors); the number of registrations appearing in more than \val{5} occurrences (re-used marks or normalization failures); and the count of occurrences dated in the future or before the source's stated coverage window.

\section{Results}
\label{sec:results}

Results are reported against the design goals. All figures are regenerable from the API calls in Appendix~\ref{app:repro} against the deposited snapshot \url{https://doi.org/10.5281/zenodo.22303316}.

\subsection{DG1: composition and provenance}
\label{sec:results:composition}

\begin{figure}[htbp]
\centering
\includegraphics[width=0.92\linewidth]{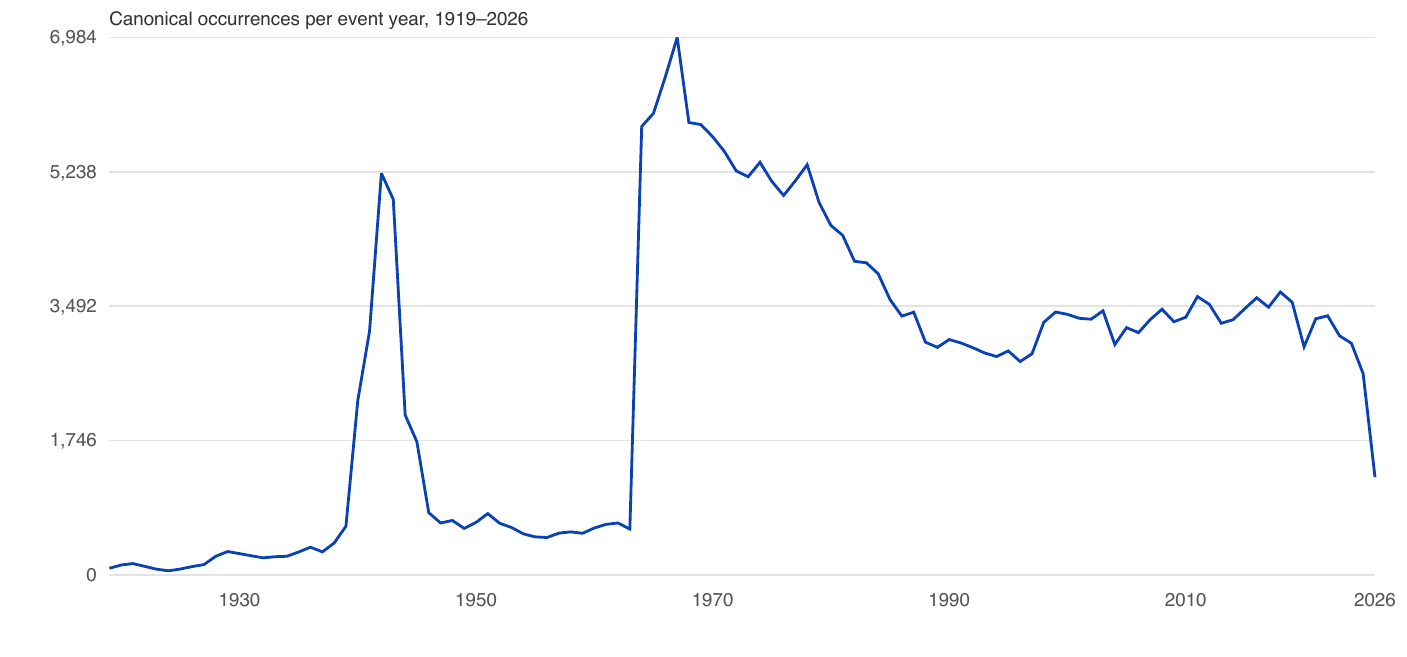}
\caption{Canonical occurrences per event year. The series is plotted from \val{1919}; the \val{36} occurrences dated \val{1902}--\val{1918} are omitted from the plot, not from the corpus. The series is a record of reporting as much as of aviation: the rise through the 1960s is the NTSB historical archive entering the corpus, and the decline through the 1980s follows the end of the 1962--1981 export. The 2007--2008 handover to the current NTSB feed leaves the annual total flat (\val{3{,}320} in 2007, \val{3{,}454} in 2008); its effect is on attribution (Table~\ref{tab:composition}), not on this series.}
\label{fig:annual}
\end{figure}

The corpus holds \val{279{,}485} occurrences assembled from \val{290{,}957} source records, of which \val{96.15}\,\% derive from a single source, \val{3.61}\,\% from two, \val{0.21}\,\% from three and \val{0.02}\,\% from four or more; the maximum for one occurrence is five, reached by \val{2} occurrences. That distribution bounds what the linkage can be said to achieve: only \val{10{,}757} occurrences (\val{3.85}\,\%) are attested by more than one authority, so every cross-source comparison in this paper rests on that subset rather than on the whole, and the remaining \val{268{,}728} single-sourced occurrences cannot be cross-checked within the corpus at all. Table~\ref{tab:composition} gives counts by decade, and Figure~\ref{fig:annual} the annual series. The corpus carries \val{131{,}882} non-empty narratives that resolve to a canonical occurrence, drawn from \val{122} of the \val{128} source codes and attached to \val{128{,}483} occurrences (\val{46.0}\,\%). The two counts differ because several narratives can attach to one event, and quoting the narrative count against the occurrence total would overstate coverage by about a point. The narrative store holds \val{164{,}068} records in all; the residual \val{31{,}933} were fetched for source records that no occurrence in this snapshot claims (typically a native identifier the ingest saw but the occurrence loader did not), and a further \val{253} resolve but carry no text. They are counted here rather than dropped silently, and \texttt{narrative\_index.csv.gz} in the snapshot lets any of these figures be recomputed under a different definition. The \texttt{region} column is empty for every row, so the geographic breakdown below is derived from \texttt{country\_iso}, which is itself absent on \val{25.2}\,\% of occurrences.

\begin{table}[htbp]
\centering\small
\caption{Canonical occurrences by decade of event date and state code
(\texttt{country\_iso}, mostly but not always the state of occurrence; see
Table~\ref{tab:variables}). The breakdown is by \texttt{country\_iso} rather than by ICAO region, because
\texttt{region} is empty for every occurrence; ``unknown'' is shown as its own
column rather than distributed, since it is the largest column in the two most
recent decades. Rows sum to \val{252{,}580}; the remaining \val{26{,}905}
occurrences fall before 1950, giving \val{279{,}485}. The ``United States''
column is a severe undercount from 2008 onward: the current NTSB feed does not
populate \texttt{country\_iso}, so its records fall into ``unknown''.}
\label{tab:composition}
\begin{tabular}{lrrrrr}
\toprule
Decade & United States & Other states & Unknown & Total & Multi-source (\%) \\
\midrule
1950--1959 & 423 & 1{,}839 & 3{,}670 & 5{,}932 & 13.1 \\
1960--1969 & 33{,}159 & 2{,}970 & 3{,}428 & 39{,}557 & 3.1 \\
1970--1979 & 45{,}305 & 3{,}604 & 3{,}420 & 52{,}329 & 3.7 \\
1980--1989 & 31{,}843 & 2{,}712 & 2{,}787 & 37{,}342 & 4.0 \\
1990--1999 & 22{,}189 & 4{,}791 & 3{,}030 & 30{,}010 & 3.7 \\
2000--2009 & 15{,}048 & 11{,}292 & 6{,}563 & 32{,}903 & 3.1 \\
2010--2019 & 536 & 15{,}870 & 18{,}432 & 34{,}838 & 3.1 \\
2020--2026 & 336 & 8{,}061 & 11{,}272 & 19{,}669 & 2.6 \\
\midrule
Total & 148{,}839 & 51{,}139 & 52{,}602 & 252{,}580 & 3.6 \\
\bottomrule
\end{tabular}
\end{table}

The shape of this table is the corpus's most consequential property and it is
invisible in the headline count. United States records dominate the middle
decades (\val{45{,}305} of \val{52{,}329} occurrences in the 1970s) and
then all but vanish: \val{536} of \val{34{,}838} in the 2010s. That collapse
is a defect of attribution in this snapshot rather than a change in the sources
or in the world. The two large historical NTSB exports, which end in 1981 and
2007, carry a state code; the current NTSB feed, which takes over in 2008 at
comparable volume (\val{1{,}775} occurrences in 2008 against the \val{1{,}825} attributed to the United States in 2007), does not. Its
normalizer never populates \texttt{country\_iso}, so \val{29{,}139} of the
\val{29{,}148} occurrences whose primary source is that feed carry no state at
all; the \val{9} exceptions took their code from a second contributing source
(Canada seven times, each from a merged TSB record) or from the registration
prefix (France and Turkey once each, states of registry).

The corpus does not assign these records to the United States, and must not:
the NTSB participates in foreign investigations, and seven of the nine records
that did receive a code take it from a Canadian TSB record. Their order of magnitude can
still be stated. \val{28{,}992} of them carry coordinates and \val{22{,}420} of
those fall inside a rectangle drawn around the contiguous United States, a box and not a border, which admits southern Canada and northern Mexico and
excludes Alaska, Hawaii and the territories. That figure gives the scale of the
missing attribution; it is not a count and is not used as one. The defect is
known, is uncorrected in this snapshot, and will be repaired when the corpus is
next recut.

What replaces the attributed records is in any case not other states so much as
\emph{unattributed} ones: \texttt{country\_iso} is missing on \val{52.9}\,\% of
2010s occurrences and \val{57.3}\,\% of those since 2020. Of the
\val{29{,}704} occurrences dated 2010 or later with no state code,
\val{25{,}664} (\val{86.4}\,\%) have the current NTSB feed as their primary
source and \val{4{,}009} (\val{13.5}\,\%) come from the global compilations,
whose country fields the normalizer does not read (B3A publishes a country name,
the ASN gives the country inside its location string). Any time series cut by
country from this corpus is therefore measuring source composition at least as
much as it measures aviation, and the decades are not comparable to one another
without saying so.

\subsection{DG2: linkage error}
\label{sec:results:linkage}

The linkage produces no weight distribution to threshold and no calibration
curve to plot. Most of it is a key ladder, which records a decision per pair
and no score at all; the scored cohort carries one, but it is a sum of fixed
rule bonuses taking eleven discrete values, so plotting it would describe the
rule rather than the data. What can be reported is the error rate of the band
that ships, measured on adjudicated samples, and the sampling frame of
each.

\begin{table}[htbp]
\centering\small
\caption{Hand-adjudicated linkage samples. Each row is an independent sample
with its own sampling frame; the frames are not nested and the rates are not
poolable. Samples~3 and~4 both come from the automatically-merging band:
sample~3 measures it as it stood before the two guards it produced,
sample~4 is a fresh draw from the band as it ships. Only sample~4 was drawn
by a recorded procedure (a uniform random sample of \val{60} of the
\val{3{,}441} applied pairs, seed \val{20260901}, listed in
\texttt{adjudicated\_pairs.csv}), and it is the only one whose selection a
reader can repeat. The frames are not nested but they are not disjoint either:
the file holds \val{200} rows and \val{198} distinct pairs, because two pairs
drawn into sample~4 had already been adjudicated in sample~1
(\texttt{src:aviation-safety:45944}\,/\,\texttt{reg:DAVFB:1938-10-01} and
\texttt{src:aviation-safety:44220}\,/\,\texttt{reg:YV84C:1985-05-28}, judged
duplicate both times). The sample~4 draw was uniform over the applied band and
was not screened against earlier samples.}
\label{tab:linkage}
\begin{tabular}{llrrrl}
\toprule
Sample & Frame & $n$ & Duplicates & Errors & Outcome \\
\midrule
1 & scoring band, as it then stood & 40 & 40 & 0 & frame too narrow \\
2 & mass fleet, small toll & 30 & 26 & 4 & cell excluded \\
2b & country agreement only & 10 & 10 & 0 & held for review \\
3 & merging band, before its guards & 60 & 58 & 1 & two guards added \\
4 & merging band, as it ships & 60 & 60 & 0 & bound, not proof \\
\bottomrule
\end{tabular}
\end{table}

\begin{figure}[htbp]
\centering
\includegraphics[width=0.92\linewidth]{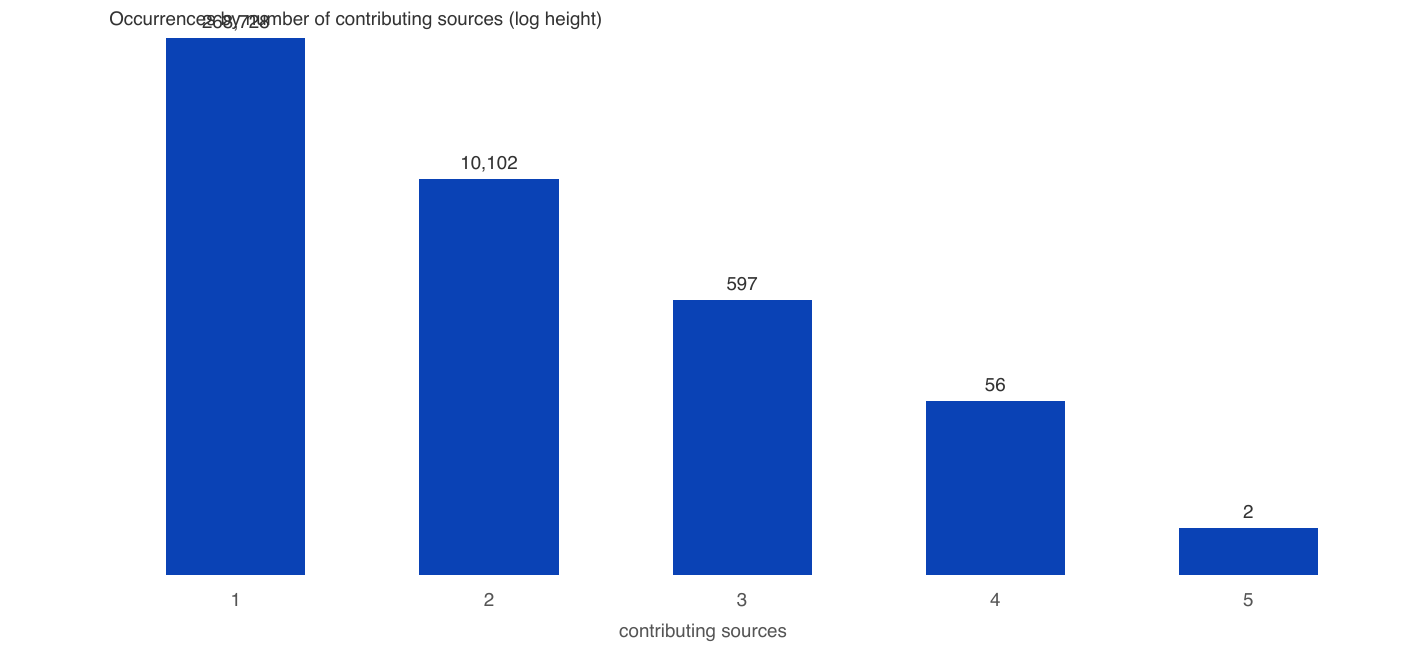}
\caption{Occurrences by number of contributing sources, log height. There is no match-weight histogram to show in its place: linkage is deterministic for all but the scored cohort, and that cohort's score takes eleven discrete values fixed by rule, so its histogram would show the rule and not the data. The multi-source bars are dominated by ladder joins: of the \val{10{,}757} occurrences with more than one source, only the \val{3{,}441} rows the scored band applies come from the mechanism whose error is measured in Section~\ref{sec:results:linkage}.}
\label{fig:weights}
\end{figure}

\begin{figure}[htbp]
\centering
\includegraphics[width=0.92\linewidth]{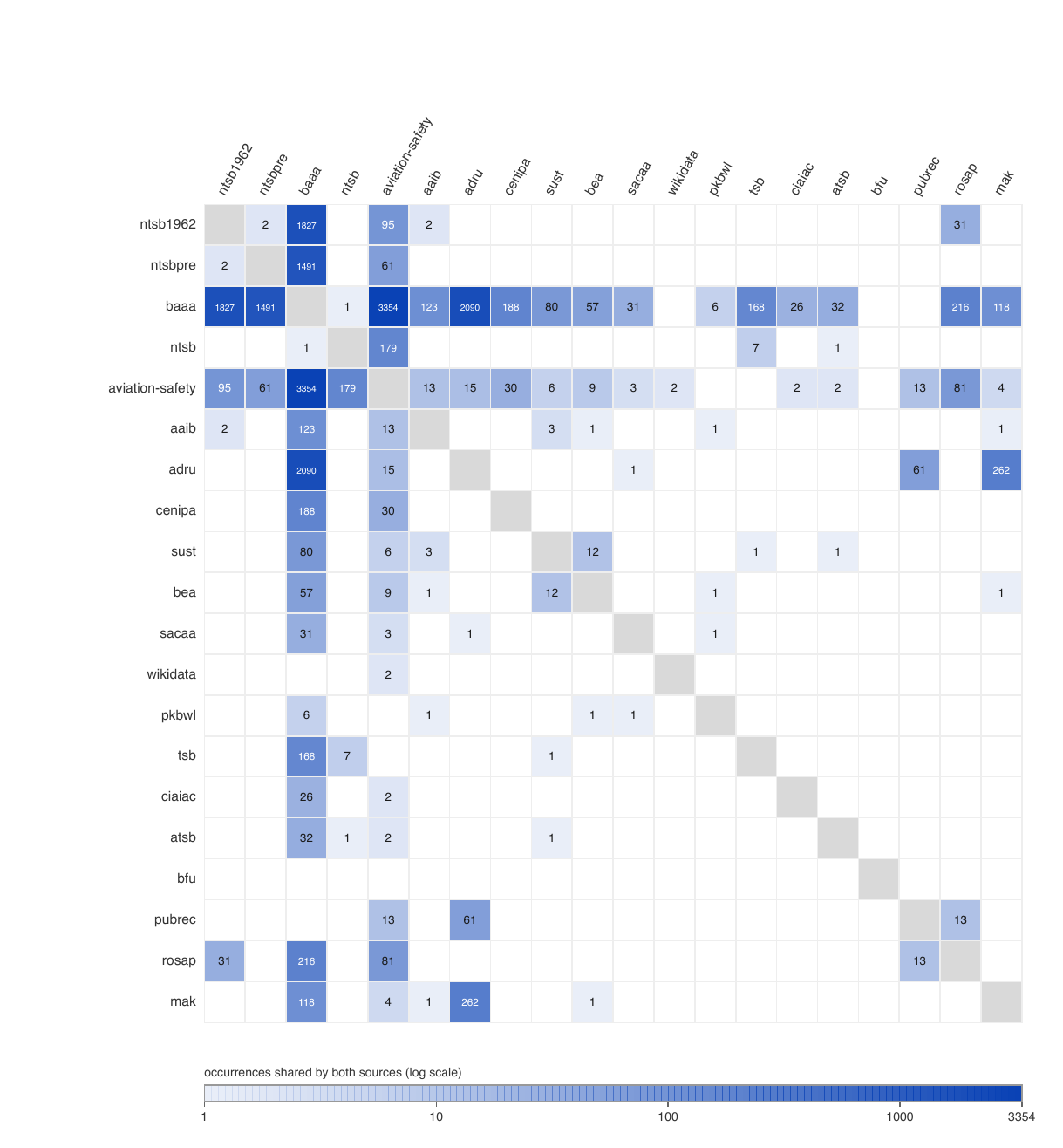}
\caption{Per-source overlap for the twenty largest source codes, ordered by contribution. Cell $(i,j)$ counts occurrences carrying records from both sources; the diagonal is each source's own total and is shown grey. Every non-zero off-diagonal cell is labelled with its count; colour is on a log scale read from the bar below, the largest off-diagonal cell being \val{3{,}354}, between the Aviation Safety Network and the Bureau of Aircraft Accidents Archives.}
\label{fig:overlap}
\end{figure}

Sample~1 is reported because it misled. Forty
duplicates and no errors reads as settled, but in all forty the deciding signal
was the operator, and the sample contained no pair held together by country
alone, so it carried no information about a stratum it never touched. Its
own interval said as much: zero errors in forty bounds the error rate below
roughly \val{7.5}\,\% by the rule of three, which over the full band admits
several hundred wrong merges.

Sample~2 was drawn deliberately from the strata sample~1 had missed and found
four genuine false merges in thirty. That cell, a mass-built airframe of a mass operator with a small death toll, is now excluded from automatic
merging.

Sample~3 was the first drawn from the band that merges automatically. It
found fifty-eight duplicates, one false merge and one pair that could not be settled. Both the error and the undecidable pair became
guards, described in Section~\ref{sec:pipeline:proposals}, and both were added
before any merge reached the corpus.

One error in \val{59} confident verdicts is \val{1.7}\,\%, and at that sample
size a single error constrains the rate only loosely: the exact binomial
(Clopper--Pearson) \val{95}\,\% interval runs from \val{0.04}\,\% to
\val{9.1}\,\%. Applied to the \val{3{,}441} pairs the band merges, the point
estimate is of order \val{58} false merges and the interval admits anywhere
from \val{1} to \val{313}. That figure is reported here as history: it describes the band as it stood when sample~3 was drawn, before the guards, and
it is superseded for the shipping band by sample~4 below. It is a point
estimate on \val{59} verdicts and never a bound: sixty adjudications cannot
exclude an error rate several times the one observed.

Both problematic pairs are now demoted to the review queue by the guards
they produced: in the deposited merge table each carries status
\texttt{pending} at score \val{0.20}, with evidence
\texttt{model-number-conflict} and \texttt{collision} respectively. Of the
sixty pairs of sample~3, \val{56} are still merged automatically and all
\val{56} were adjudicated duplicates, but that is in-sample by
construction, because the guards were derived from the very two pairs they now
catch, and it cannot stand as a rate for the current band. Four pairs left
the automatic band, not two, and only three of them left it because of a
guard: besides the two the guards were derived from, a merge this sample had
adjudicated as a duplicate now sits at score \val{0.50} carrying
\texttt{collision}. The fourth, also adjudicated a duplicate, carries no guard
marker at all. Its evidence string ends in \texttt{high-toll-anchor}, which is a scoring bonus and not a guard, borne by \val{1{,}322} of the
\val{3{,}441} rows the band applies, so at score \val{0.65} the pair falls below
the cut-off on the score alone. Counted in pairs, the guards withheld one
correct merge to withdraw one wrong merge and one pair that could not be
decided, a cost in recall that is visible only because these sixty pairs
happen to have been adjudicated.

Sample~4 was drawn to answer that. It is a uniform random sample of \val{60}
of the \val{3{,}441} pairs the band applies today, drawn with seed
\val{20260901} so the selection can be repeated, and adjudicated out of
sample: none of the sixty is a pair any guard was derived from. All \val{60}
were duplicates. Fifty-three were settled mechanically (the two sides carry
the same registration on the same date, in one case encoded in the absorbed
record's own key rather than in its fields, and a registration is not reissued
within a day) and the remaining seven by reading the sources' own records,
with the deciding fact recorded per pair. That split limits what the sample
establishes. Registration and date agreement is the signal the merge rule
already weighs most heavily, so for those fifty-three the adjudication confirms
the rule's own input rather than re-establishing the event independently. Only
the seven decided from the sources' records bring evidence the rule did not
use.

Zero errors in sixty does not mean the band is clean; it bounds how unclean it
can be. The exact (Clopper--Pearson) two-sided \val{95}\,\% upper bound on
the rate is \val{5.96}\,\%, which across the \val{3{,}441} applied pairs is
\val{205} false merges; the one-sided \val{95}\,\% bound, appropriate if
only the upper direction is of interest, is \val{4.87}\,\% or \val{168}
pairs. Neither is a ceiling on the number of false merges: both are confidence
statements about the rate, and a reader who wants the two-sided bound below
\val{1}\,\% would have to adjudicate \val{368} pairs at zero errors
(\val{299} one-sided). That bound is
wider than the point estimate sample~3 gave, so these two samples together do
not demonstrate that the guards improved anything; sixty observations cannot
show that. What sample~4 establishes is narrower and was previously missing
altogether: the band as it ships has now been measured on a sample whose
selection is recorded, and it produced no error.

The figure bounds clerical
error only: the two global compilations dominating this cohort are not
independent of each other, so a duplicate that both inherited from one upstream
mistake would be adjudicated as a duplicate and counted as correct.

\subsection{DG3: harmonization outcome}
\label{sec:results:harmonization}

Table~\ref{tab:crosstab} cross-tabulates canonical \texttt{occurrence\_type} against \texttt{severity}. Cross-source class disagreement is not reported: the corpus stores one harmonized class per occurrence and does not retain the losing record's class after the priority rule has run, so the disagreement rate cannot be recovered from the published snapshot. Reporting it would require the loader to persist each contributing record's pre-merge class, which is a schema change rather than an analysis and is listed under future work in Section~\ref{sec:futurework}. \val{0.30}\,\% of \texttt{fatal} occurrences carry a canonical class other than \texttt{accident} (\val{176} of \val{58{,}473}), of which \val{169} are \texttt{combat\_loss}. Field completeness after normalization is: date \val{100}\,\%, registration \val{78.0}\,\%, \texttt{type\_icao} \val{15.6}\,\%, \texttt{country\_iso} \val{74.8}\,\%, operator \val{41.5}\,\%, \texttt{fatalities\_total} \val{52.5}\,\%.

\begin{table}[htbp]
\centering\small
\caption{Canonical occurrences by harmonized occurrence type and severity. The
two fields are not independent axes of one classification: \texttt{severity}
carries a \texttt{serious\_incident} level that \texttt{occurrence\_type} does not, and every record holding it is typed \texttt{accident}; that level marks an ATSB serious injury, not the Annex~13 class (Section~\ref{sec:pipeline:mapping}).}
\label{tab:crosstab}
\footnotesize
\setlength{\tabcolsep}{4pt}
\begin{tabular}{lrrrrrrr}
\toprule
\texttt{occurrence\_type} & \texttt{fatal} & \texttt{hull\_loss} & \texttt{serious\_inc.} & \texttt{incident} & \texttt{minor} & \texttt{unknown} & Total \\
\midrule
\texttt{accident} & 58{,}297 & 9{,}728 & 69 & 196 & 672 & 208{,}195 & 277{,}157 \\
\texttt{incident} & 7 & 5 & 0 & 0 & 0 & 2{,}061 & 2{,}073 \\
\texttt{combat\_loss} & 169 & 10 & 0 & 0 & 0 & 76 & 255 \\
\midrule
Total & 58{,}473 & 9{,}743 & 69 & 196 & 672 & 210{,}332 & 279{,}485 \\
\bottomrule
\end{tabular}
\end{table}

Three cells deserve comment. \texttt{unknown} severity covers
\val{75.3}\,\% of the corpus, so the table is mostly a statement about what the
sources did not record; a study treating absent severity as non-fatal would
misclassify three quarters of the data. The \val{69} \texttt{serious\_incident} records typed \texttt{accident} are ATSB serious-injury accidents under a misleading severity label (Section~\ref{sec:pipeline:mapping}); no source's Annex~13 serious-incident class survives anywhere in the corpus. And \val{169} of the \val{255}
\texttt{combat\_loss} records are fatal; these are shoot-downs and war
losses, retained for provenance and excluded from every rate reported here.

\section{Discussion}
\label{sec:discussion}

\subsection{What the multi-source share means}
\label{sec:results:multisource}

\val{10{,}757} occurrences (\val{3.85}\,\%) carry records from more than one
source. That share is the corpus's own cross-check surface, the only place where two authorities can be observed describing one event, and it is the
quantity the merge work of Section~\ref{sec:pipeline:merges} moves. Before the
adjudicated merges were applied the same corpus reported \val{7{,}558}
multi-source occurrences (\val{2.67}\,\% of a then-larger total); the merges
raised the count by \val{42}\,\%, not by finding new records but by
recognising that pairs already present described one event. Like the
recovery-pass figures of Section~\ref{sec:pipeline:merges}, this is a
run-time measurement of a state the deposit no longer holds: the snapshot
carries the corpus after those merges, so \val{7{,}558} cannot be recomputed
from it and is reported as measured rather than as checkable.

In a corpus where \val{96}\,\% of occurrences come from a single source,
the linkage leaves that majority as it is. It changes the size of the minority
on which every cross-source claim in this paper and its companions rests, and
it changed it by nearly half.

The multi-source share stands in for the external coverage estimate that
Section~\ref{sec:validation:coverage} explains cannot be computed. It measures
the pipeline, not the world.

\subsection{What the corpus is and is not}

The corpus is a harmonized, provenance-preserving aggregation of published official records. The design goals commit it to breadth, to a measured deduplication error, to a small common vocabulary and to reproducibility; they do not commit it to completeness, which no aggregation of published records can achieve, or to correctness beyond that of its sources. The validation results should be read in that light. The band that merges automatically was measured by a recorded random draw of \val{60} of the \val{3{,}441} pairs it applies, which found no error, bounding rather than disproving a rate: the exact two-sided \val{95}\,\% upper bound is \val{6.0}\,\%, a confidence statement about the rate and not a ceiling on the count. Recall is not measured at all, and that omission is the more serious of the two, because the ladder errs toward leaving records apart, so the duplicates it fails to join stay in the corpus as separate occurrences and inflate counts. Only the measured error can be placed on a scale: \val{205} pairs at that bound, \val{0.07}\,\% of \val{279{,}485} occurrences, far below the between-source coverage differences visible in Figure~\ref{fig:overlap}. No comparable statement is possible for the unmeasured one, and nothing in this paper should be read as bounding it. The measured side can be tightened by users who need it: \texttt{occurrence\_merges.csv} publishes every proposed pair with its status, its rule-derived score and the evidence string that produced it, so the \val{627} pairs still in the review queue can be adjudicated further and the applied band re-cut, resolving identifiers through \texttt{event\_sources.csv.gz}, since \val{85} rows name a survivor the ladder has since re-keyed (Section~\ref{sec:pipeline:merges}). What is not published is a calibrated weight, because none is estimated.

\subsection{Alternative explanations for the validation results}

Several artefacts could produce the numbers in Section~\ref{sec:results} without the pipeline working as intended. \emph{An error rate that flatters the band.} Sample~1 measured 40 of 40 and meant less than it appeared, because its frame excluded the strata where errors live; the guard against this is not a wider sample but a stated frame, and Table~\ref{tab:linkage} gives the frame of each. \emph{Adjudication agreeing with the corpus by construction.} The two global compilations dominating the unkeyable cohort read the same official reports, so a duplicate both inherited from one upstream transcription error is adjudicated as a duplicate and counted correct; the reported rate therefore bounds clerical error and not source error, and no design available here separates them. \emph{Under-merging read as precision.} The ladder errs toward leaving records apart, so a corpus that merged nothing would report no false merges at all, which is why the multi-source share of Section~\ref{sec:results:multisource} is reported beside the error rate: it is the quantity that falls if the linkage becomes timid. \emph{Ingestion lag mistaken for under-reporting.} Low apparent coverage for a state may mean an authority publishes in a form the adapters do not yet parse; the per-source coverage window in Table~S1 separates the two.

\subsection{Comparison with prior resources}

Relative to ASN and B3A, the corpus is broader in source and carries their editorial judgement with them: it holds the national authorities' records that neither compilation rewrites, and it also contains the events they record that no ingested authority describes, because both compilations are ingested as sources rather than filtered against an official list. It does not claim to be broader in aircraft class, since the ASN Wikibase it ingests already extends to light aircraft and rotorcraft. On this snapshot \val{58{,}425} occurrences (\val{20.9}\,\%) have no contributing source other than the four compilations (\val{21{,}217} carry only ASN, \val{31{,}558} only B3A, \val{55{,}770} nothing beyond those two), and these are what give the corpus its coverage outside investigated events. The deposit cannot separate compilation entries that cite an official report from those built on press coverage; a user who needs authority-level evidence should treat compilation-only occurrences as a stratum and filter on \texttt{event\_sources.csv.gz}, as Section~\ref{sec:validation:coverage} already does for the coverage argument. Relative to the national databases, the corpus adds cross-reference and a common vocabulary at the cost of the field depth each authority provides natively; the source-specific fields are \emph{not} carried over (the API's \texttt{facts\{\}} object is a fixed summary of canonical values, not a per-source field store), so a study needing native depth must follow the URL in \texttt{event\_sources} back to the source. \citet{omrani2024assessment} trained injury-severity models on NTSB, TSB and ATSB records separately and tested whether the features the three share could support one common model. A cross-authority model of that kind has to deal with coding differences between the authorities, and this corpus does not resolve them: the allow-list of Section~\ref{sec:pipeline:mapping} collapses every unrecognized national class into \texttt{accident} and keeps no record that it did, so the coding differences are erased rather than reconciled, and \texttt{occurrence\_type} must not be used as the outcome of a cross-authority comparison. What the corpus does add for that purpose is the source axis: the overlap matrix identifies the events several authorities described, and \texttt{event\_sources} names which, so a study of coding disagreement can be run on the raw records behind those occurrences; retaining the pre-merge class is listed under future work in Section~\ref{sec:futurework}. The corpus follows the descriptor precedent set for aviation by the OpenSky datasets \citep{strohmeier2021crowdsourced} in pairing a live API with frozen, DOI-referenced snapshots, and adopts the datasheet structure of \citet{gebru2021datasheets} in Appendix~\ref{app:datasheet}. The author's earlier open aggregate of the FAA wildlife-strike database \citep{kolomiiets2026wildlife} used the same versioned-release pattern for a single source; the present work extends it to many.

\subsection{Intended uses and cautions}

The corpus is suited to questions in which source attribution matters or in which breadth across jurisdictions outweighs field depth: cross-national comparison of occurrence mix, studies of disagreement between authorities, text mining over narratives with per-source licensing respected, and as a linkage target for other datasets keyed on registration and date. It is not suited, on its own, to rate estimation, because it carries no exposure denominators \citep{oster2013analyzing}, or to trend estimation across the coverage window of any source without conditioning on that window, because Figure~\ref{fig:annual} shows that the series is dominated by when authorities began publishing. Users should stratify by source and occurrence class before pooling, should treat single-source occurrences from sources with no overlap as unvalidated, and should treat narratives under \texttt{excerpt} policy as pointers to the original rather than as text data.

Three aggregations should not be attempted on this snapshot at all; each follows from a property stated elsewhere in this paper. \emph{Do not sum \texttt{fatalities\_total} over the corpus.} The unit is one aircraft's involvement and not the event (Section~\ref{sec:data:unit}), so where an authority reports a collision's combined toll on each airframe's record the toll is counted once per occurrence; the sum is in addition conditional on the \val{52.5}\,\% of occurrences that carry the field. The deposit permits neither the double count nor the missing half to be corrected, so a corpus-wide death toll is not a quantity this snapshot can produce. \emph{Do not cut a country time series across 2008.} The current NTSB feed does not populate \texttt{country\_iso}, so the ``United States'' column of Table~\ref{tab:composition} collapses at that handover for a reason that has nothing to do with aviation (Section~\ref{sec:results:composition}). The break is a defect of this snapshot, and any per-state series spanning it measures the defect. \emph{Do not compute figures by aircraft type.} \texttt{type\_icao} resolves on \val{15.6}\,\% of occurrences, and the residue is not missing at random: resolution runs from \val{25.7}\,\% on the largest NTSB export down to \val{1.5}\,\% on B3A and \val{0} on Wikidata, so a per-type count is a count of the sources whose model strings the lookup table happens to cover.

\subsection{Limitations}
\label{sec:limitations}

\subsubsection{One source is redistributed without a licence}
\label{sec:limitations:licence}

Provenance is this corpus's claim, so a gap in it is reported first, and the
gap is wider than a count of non-empty \texttt{license} fields suggests. The
field is free text, and across the \val{128} contributing codes it falls into
four classes. \val{15} codes (\val{132{,}541} source records) name
recognisable terms: CC~BY~4.0, CC0, the UK Open Government Licence,
Licence~Ouverte/Etalab~2.0, United States public domain, the SKYbrary licence
agreement, and the Government of Canada non-commercial reproduction terms.
\val{107} codes (\val{127{,}436} records) carry nothing but the name of the
publishing organisation, among them \texttt{ntsbpre} (\val{60{,}400}) and
\texttt{baaa} (\val{41{,}317}); the corpus asserts a rights-holder for these,
not a grant. \val{5} codes (\val{6{,}183} records) carry a descriptive note
about the material rather than terms. And one records an explicit absence.
The Aviation Safety Network publishes no usable terms: its footer copyright
link resolves to a 404, leaving a liability disclaimer and a copyright notice,
and the records this corpus ingests come from the user-contributed ASN Wikibase
rather than the curated ASN database. The registry entry therefore records the
copyright holder and the absence of a grant instead of asserting terms that
were never given. What this corpus asserts, then, is that it has recorded
whatever each source published about reuse; it does not assert that
\val{127} of \val{128} sources have granted redistribution rights, and the
\val{107} bare organisation names should be read as an open question per
source rather than as a permission.

The exposure is not small: ASN supplies
\val{24{,}797} source records (\val{8.5}\,\%) and is the primary source of
\val{21{,}242} occurrences (\val{7.6}\,\%), so its field values are
redistributed in the snapshot. Facts about an accident (its date, registration, location and toll) are not themselves protectable, and every
ASN-derived record carries a link back to the ASN page it came from; but a
substantial extraction from a compiled database engages the European
\emph{sui generis} database right independently of copyright in the facts, and
this paper does not claim that question is settled. A permission request to the
Flight Safety Foundation is outstanding. Users who need to exclude the source
can: \texttt{event\_sources.csv.gz} carries one row per contributing source, so
dropping \texttt{source = 'aviation-safety'} and the occurrences left with no
other source is a two-line filter. Reproduced on this snapshot it removes the
\val{24{,}797} ASN rows and \val{21{,}217} ASN-only occurrences, leaving
\val{258{,}268}. The filter does not make the remainder free of ASN, and it is
not offered as such: exactly \val{25} surviving occurrences still carry
\texttt{aviation-safety} as their primary source, all of them ASN and SKYbrary
together with the rank tie broken by first arrival, and a further \val{3{,}580}
carry an ASN row beside a higher-ranked source, so some of their null fields may
have been filled from it. \texttt{events.jsonl.gz} holds no per-field
provenance, so that filling cannot be undone from the deposit.

\subsubsection{Fields that are mostly empty}
\label{sec:limitations:fields}

{\sloppy
The schema is wider than the data. Any cut along a sparse field silently
changes its own denominator, so the completeness of every variable used in this
paper is given here rather than in a supplement, measured on the deposited
snapshot: \texttt{event\_date} and
\texttt{report\_url} \val{100}\,\%,
\texttt{registration} \val{78.0}\,\%, \texttt{country\_iso} \val{74.8}\,\%,
\texttt{phase} \val{70.1}\,\%, \texttt{category} \val{58.9}\,\%,
\texttt{fatalities\_total} \val{52.5}\,\%, \texttt{operator} \val{41.5}\,\%,
\texttt{occupants} \val{28.2}\,\%, \texttt{lat} \val{16.9}\,\%,
\texttt{type\_icao} \val{15.6}\,\%, \texttt{flight\_nature} \val{14.8}\,\% and
\texttt{manufacturer} \val{0.4}\,\%. Two fields named elsewhere in this paper are returned by the API but are not columns of \texttt{events.jsonl}, so the snapshot supports no figure for them and none is quoted: \texttt{cause\_summary} and \texttt{region}. The latter is
null on every row the API returns and is discussed immediately below.\par}

Three of these deserve naming rather than tabulating. \texttt{region} is empty
for every occurrence, so no regional analysis is possible from this corpus
without inferring region from \texttt{country\_iso}, which is itself missing on
a quarter of records and missing most on the recent decades, there because
the current NTSB feed's normalizer leaves \texttt{country\_iso} null, a defect
of this snapshot rather than of the source. \texttt{type\_icao}
at \val{15.6}\,\% means a per-type rate computed from this corpus describes the
sixth of it that carries a designator, not the fleet. And
\texttt{fatalities\_total} at \val{52.5}\,\% means that a sum of deaths over the
corpus is a sum over the half of records that report them; the figure quoted
anywhere in this paper is that conditional sum and not an estimate of deaths.

\subsubsection{Consistency findings}

Three of the integrity checks return non-zero. \val{4{,}945} occurrences sit at latitude and longitude both exactly zero, and \val{6{,}312} lie within the $0.05^{\circ}$ radius the key ladder treats as unknown,
which is not a position in the Gulf of Guinea but a source's way of writing
``unknown''; the key ladder already refuses to link on such a position, but a
map built from this corpus without filtering will draw them. \val{176}
occurrences are fatal but not typed \texttt{accident}, of which \val{169} are
\texttt{combat\_loss} and \val{7} \texttt{incident}. \val{90} registrations
appear in more than five distinct occurrences, which is plausible for a
long-lived training airframe and is also what a normalisation error looks like;
they are flagged, not resolved. Future-dated events: \val{0}.

\subsubsection{One state, four source codes}

Four NTSB source codes supply \val{60.8}\,\% of source records. The corpus is
therefore a United States dataset with substantial international coverage
rather than a global one with an American component, and every unqualified
aggregate in it inherits that shape. This is not correctable by weighting: the
absent records are absent because the authorities that would have filed them
publish nothing machine-readable, and a weight cannot recover an event nobody
wrote down.

The most important limitation is coverage bias inherited from the sources: events that no source published in any form are absent, and \citet{icao2025wp330} tie non-publication of final reports to capacity shortfalls and uneven compliance with Annex~13 in several states. Their figure bounds report availability rather than corpus coverage (an event without a final report may still enter through a preliminary report, a reporting system or a compilation), but its dependence on state capacity is the relevant warning either way. The corpus cannot estimate this from within at all; Section~\ref{sec:validation:coverage} explains why the natural external check is unavailable. Second, the linkage relies on registration and date; events where the registration is absent from all sources, or where an authority reports only a month, depend on the weaker blocking keys and have lower recall, which is concentrated in older records and in sources with sparse structured fields. Third, the mapping from native classes to three harmonized levels loses information and, for sources with a single undifferentiated class, imposes a severity-based rule that Annex~13 does not sanction; the native class is not retained in the published objects (neither the API's \texttt{facts\{\}} nor \texttt{events.jsonl.gz} carries it), so the canonical class must be read as the output of that default rule and the native class recovered from the source. Fourth, the samples were adjudicated with a large language model under the author's direction (see Assisted work), working from the sources' own published records rather than independently of the project; adjudication by the party that built the linkage is not independent, and the reported rates should be read with that in mind. Fifth, narratives are multilingual and are not translated, and the corpus makes no attempt to remove names of individuals that authorities themselves publish. Finally, the corpus is a moving target: sources are added and re-ingested, thresholds may be re-tuned, and the figures in this paper describe the snapshot deposited as version \val{2026-09-04} (DOI 10.5281/zenodo.22303316) and no other.

\subsection{Future work}
\label{sec:futurework}

Three extensions follow naturally. Exposure denominators from fleet and schedule data would convert counts into rates and are the subject of a companion paper in preparation. Field-level disagreement between authorities, of which the class disagreement in Section~\ref{sec:results:harmonization} is one instance, is analyzed in a second companion paper on cross-source disagreement, in preparation. Structured extraction from narratives \citep{tanguy2016natural} could recover the source-specific detail the harmonized schema drops into the ADREP vocabulary; phase of flight in particular is carried today as the source's own unmapped string or as an unmapped model extraction (Section~\ref{sec:pipeline:other}); a pipeline for that is evaluated separately in a companion paper on narrative text mining, in preparation. On the linkage side, replacing the point-estimate thresholds by a posterior over partitions \citep{sadinle2017bayesian} would allow downstream counts to carry linkage uncertainty. Two smaller changes would answer questions this paper has to leave open. Persisting each contributing record's pre-merge occurrence class would make the cross-source disagreement rate of Section~\ref{sec:results:harmonization} measurable instead of unrecoverable; it is a schema change, not an analysis. And recall, which is not measured here at all, admits a cheap internal lower bound: mask the registrations on a sample of the \val{10{,}757} occurrences already linked by registration and count how many the scoring pass recovers without them. That would not measure the duplicates no source keyed, but it would replace an unmeasured quantity with a measured floor.

\section{Conclusion}
\label{sec:conclusion}

The \ff{} Occurrence Corpus assembles \val{279{,}485} aviation occurrences and \val{131{,}882} attached narratives from the \val{128} source codes that contribute to it (national investigation authorities and occurrence-reporting systems, four global compilations, and one public-records channel) into one schema, with the identity and URL of one record per contributing source preserved on each occurrence and the attribution and licence terms carried on the source code it belongs to. Provenance is keyed on \texttt{(occurrence, source)}, so where one source supplies two records of the same occurrence only one survives in provenance. The construction pipeline is documented stage by stage: per-source ingestion into a staging table keyed on source and native identifier and overwritten on re-sync, normalization of dates, registrations, aircraft types and locations, a deterministic key ladder in which each record reduces to exactly one identifier so that a missing field cannot create a link, an explicit merge table for the cohort the ladder cannot reach, an authority-priority rule for canonical values, and a closed allow-list with a stated fall-through default for occurrence type and severity, which is why \texttt{accident} at \val{99.17}\,\% is a property of that default and not of aviation. Steps outside those stages, among them a duplicate pre-filter on three sources and narrative values extracted by a language model, are stated as well. On the band that merges automatically, a recorded random draw of \val{60} found no error, bounding its false-merge rate at \val{6.0}\,\% with \val{95}\,\% confidence; coverage relative to the ASN and B3A compilations is not reported, because both are ingested by the corpus and the comparison would measure a subset of it against itself; see Section~\ref{sec:validation:coverage} for why no capture--recapture estimate of the events missed by every list is offered. The corpus carries no exposure data and inherits the under-reporting of its sources, and it is offered not as a complete census of aviation occurrences but as a documented, provenance-preserving base whose reported statistics can be recomputed from the deposited snapshot on which the coverage and disagreement of the world's investigation systems can themselves be studied. A frozen snapshot, the adjudicated linkage sample and the source registry accompany the live API so that every table in this paper can be regenerated and every future release compared with it.

\section*{Data and code availability}
\phantomsection\label{sec:availability}
\addcontentsline{toc}{section}{Data and code availability}

The corpus is served by the \site{} data API, base URL \url{https://himaxym.com/api/v1/data}, documented by the OpenAPI specification at \url{https://himaxym.com/api/v1/data/openapi.json}. The endpoints used in this paper are \apiep{GET /sources} (registry), \apiep{GET /events} (paginated occurrences with filters), \apiep{GET /events/\{id\}} (occurrence with all contributing sources and narratives), \apiep{GET /narratives/\{source\}/\{id\}} and \apiep{GET /aircraft/\{family\}/safety}. The \apiep{GET /sources} and \apiep{GET /events} endpoints answer without credentials at a low rate; sustained use requires a free API key (1{,}000 requests per day, 60 per minute) or a Pro key (20{,}000 per day); full narrative text under \texttt{full} policy is available on the Pro tier only. Structured fields are redistributed under the licence stated per source code in the \texttt{license} field of the registry, returned inline on each element of \texttt{sources[]} by \apiep{GET /events/\{id\}}; narrative text is redistributed only under the source's \texttt{policy}, and users must reproduce the \texttt{attribution} string. The frozen snapshot described in this paper, comprising the occurrence and source tables, the registry, the narrative index and the adjudicated linkage sample with its verdicts, is deposited at Zenodo under concept DOI 10.5281/zenodo.22228451, which always resolves to the most recent version; the exact snapshot from which every number reported here is computed is version DOI 10.5281/zenodo.22303316. The deposit also carries the two scripts that regenerate the manuscript's descriptives and figures from the deposited files, so every reported number other than the run-time figures marked as such can be recomputed without access to the live database. The ingestion, normalization and linkage modules are not in the deposit: they are part of a larger application that is not public, and are specified in the construction and linkage sections above rather than released. The deposited merge table records every decision of the scoring pass and \texttt{event\_sources.csv.gz} shows which sources each occurrence joins, but the records hidden by the pre-filter of Section~\ref{sec:pipeline:other} and the same-source siblings dropped from provenance are recorded nowhere in the deposit. For the corpus to be rebuildable from source records, a future release would need to ship either the scoring module itself as a standalone executable with its dictionaries (the placeholder word list, the type--token fold table, the collision markers and the boilerplate patterns), or a complete numeric specification of it, every rule bonus, the high-toll anchor's definition and trigger toll, the guard mechanics and the band cut-off, with worked test pairs whose scores and statuses can be checked against \texttt{occurrence\_merges.csv}. The deposited files are released under CC BY 4.0 except where the source's own terms forbid it, and three sources are carved out by name: the Transportation Safety Board of Canada (\val{1{,}317} records, Crown copyright reproduced under the Government of Canada \emph{non-commercial} reproduction terms), SKYbrary (\val{464} records, under a bespoke EUROCONTROL agreement) and the Aviation Safety Network (\val{24{,}797} records, which grants nothing; see Section~\ref{sec:limitations:licence}). Rows from those three are redistributed under their own terms and are not relicensed here. All structured fields remain subject to the per-source attribution recorded in Table~S1.

\section*{Author contributions and competing interests}
\addcontentsline{toc}{section}{Author contributions and competing interests}

D.K.\ designed and built the corpus and pipeline, designed and executed the validation, and wrote the paper. The author operates \site{}, the website and API through which the corpus is published, and has no other competing interest to declare.

\paragraph{Assisted work.} The pipeline code, the adjudication of the five linkage samples of Table~\ref{tab:linkage} and successive drafts of this paper were produced with a large language model under the author's direction. A language model also produced data: the narrative extractions that supply \texttt{occupants} and, where no source gives them, fatality counts, \texttt{phase} and \texttt{category} (Section~\ref{sec:pipeline:other}), and the \texttt{article} write-ups the API serves; the extracted values carry no flag in the snapshot. Adjudication verdicts were reached from the sources' own published records, with the basis for each verdict recorded per pair in the deposited \texttt{adjudicated\_pairs.csv}; for samples~1--3 that basis is a category rather than a citation of the specific record, while sample~4 names the deciding fact. No verdict rests on the model's recollection. The author is accountable for every number and every claim here and has checked each against the deposited snapshot and the pipeline source. Where a check was not performed it is marked as such in the text rather than left to inference.

\appendix

\section{Reproducibility protocol}
\label{app:repro}

The following steps regenerate every table and figure from the API and the snapshot. Three levels of reproduction must be kept apart, and only the first is supported here. \emph{(i)~Re-deriving this paper's statistics from the snapshot} is supported for every number reported above except the run-time figures marked as such: \texttt{describe.mjs} regenerates the headline descriptives in \texttt{descriptives.json}, and the remaining counts are filters over the deposited files that the deposit does not script. \emph{(ii)~Rebuilding the corpus from source records} is not supported: the ingestion, normalization and scoring modules are closed (Section~\ref{sec:availability}), the numeric specification of the scoring pass is not published, the source records' own field values are not deposited, and the records hidden by the pre-filter of Section~\ref{sec:pipeline:other} are not deposited at all. \emph{(iii)~Reversing a merge} is possible only as far as naming it: \texttt{event\_sources.csv.gz} carries \texttt{(occurrence\_id, source, native\_id, url)} and no field values, so a reader can identify the absorbed source record and retrieve it from its source, but cannot recover the values it held in the corpus. Environment: Node.js \val{24.x} for the pipeline itself, which has no statistical dependencies --- the linkage is a key ladder and a rule-based scoring pass, so there is no model to fit, no library to pin and no seed to fix. No Python and no version pinning are required: \texttt{describe.mjs} and \texttt{make\_figures.mjs} in the deposit use only the Node standard library, and both resolve their inputs next to themselves, so they run unchanged from the flat Zenodo download (\texttt{node describe.mjs}) or from a checkout (\texttt{node deposit/describe.mjs}), and the figures are emitted as SVG and converted with \texttt{rsvg-convert}

\begin{enumerate}[leftmargin=2.2em]\sloppy
\item \textbf{Registry.} \texttt{curl -H "Authorization: Bearer \$KEY"} \url{https://himaxym.com/api/v1/data/sources} returns the registry (\texttt{code, name, license, homepage, narratives, policy}). Save as \texttt{sources.json}. Note that the registry and the corpus are not the same list: the registry is the set of codes \emph{permitted} to enter, and at the time of this snapshot two of them (\texttt{bdca}, \texttt{dinacpy}) had contributed no records yet. Table~S1 and the source axis of Figure~\ref{fig:overlap} are the \val{128} codes that actually contribute, which is \texttt{event\_sources.csv.gz} grouped by \texttt{source} --- take that as the definition, not the registry's row count.
\item \textbf{Occurrences.} Page through \texttt{GET /events?from=\val{YYYY-MM-DD}\allowbreak\&to=2026-\val{MM-DD}\allowbreak\&limit=100\allowbreak\&cursor=...} until \texttt{next\_cursor} is null, writing each \texttt{data[]} element to \texttt{events.jsonl}. The snapshot contains the identical table and needs no paging. Optional filters \texttt{country}, \texttt{type}, \texttt{family}, \texttt{operator}, \texttt{fatal=1} are not used for the full pull.
\item \textbf{Contributing sources.} For each occurrence, \texttt{GET /events/\{id\}} returns \texttt{sources[]} and \texttt{narratives[]}, each source element carrying \texttt{source}, \texttt{attribution}, \texttt{license} and \texttt{url}. To avoid \val{279{,}485} calls the snapshot ships a reduced projection of the same relation, \texttt{event\_sources.csv.gz}, with columns \texttt{(occurrence\_id,\allowbreak source,\allowbreak native\_id,\allowbreak url)} only. Attribution and licence are not repeated per row: they are properties of the source code, and the snapshot carries them once per code in \texttt{table\_S1.csv}, which is joined on \texttt{source}. A reader who needs licence terms at record level must perform that join, or call the API, where the fields are returned inline. For \texttt{ntsb}, \texttt{aviation-safety} and \texttt{wikidata} the \texttt{native\_id} is the row number of \ff{}'s own staging store rather than the source's identifier, which is carried in the URL (\val{1{,}805} rows of these three sources hold several comma-joined URLs, some of them another source's page), and the \texttt{src:} keys of those three sources use the same row number.
\item \textbf{Narratives.} The snapshot carries no narrative text --- licences differ per source and several permit excerpting only --- but it ships \texttt{narrative\_index.csv.gz}, one row per stored narrative with columns \texttt{(occurrence\_id,\allowbreak source,\allowbreak native\_id,\allowbreak slug,\allowbreak narrative\_chars,\allowbreak probable\_cause\_chars,\allowbreak indexable)}. Every narrative count in this paper is a filter over that file: \val{164{,}068} rows in all, \val{132{,}135} whose \texttt{occurrence\_id} is non-null (the narrative resolves to a canonical occurrence), and \val{131{,}882} of those with non-empty text --- the figure quoted in Section~\ref{sec:results:composition}, covering \val{128{,}483} occurrences from \val{122} source codes. Text itself is retrieved per record through \apiep{GET /narratives/\{source\}/\{id\}} under that source's \texttt{policy}.
\item \textbf{Deduplication rules (for rebuilding from source records).} Normalize registration as \texttt{regexp\_replace(\allowbreak upper(reg),\allowbreak '[\textasciicircum A-Z0-9]',\allowbreak '')}, upper-casing first and stripping no annotation, and reject the placeholder word list and any string with fewer than three letters or digits before using the result. Parse dates to a local calendar date, retaining internally whether a day, a month or only a year was resolved; that flag is not a field of the published schema. Then apply the ladder of Section~\ref{sec:pipeline:linkage} in order, taking the first rung that applies --- event signature for the three tier-aware sources, registration and date, position rounded to one decimal and date, source and native identifier --- and group records by the resulting key; the position is rounded as JavaScript's \texttt{toFixed(1)} rounds the stored double. Applied to the published fields of \texttt{events.jsonl.gz}, these rules reproduce all \val{216{,}528} \texttt{reg:} and \val{40{,}656} \texttt{geo:} identifiers and \val{21{,}959} of the \val{22{,}018} \texttt{src:} ones; the other \val{59} gained a position or registration from a merged record after keying, and the \val{283} \texttt{evt:} keys need narrative text the deposit does not carry. That checks the key function and not the grouping, which needs the source records' own fields. There is no model to fit, no threshold to sweep and no seed to fix. For the cohort that reaches the last rung, the scoring rules of Section~\ref{sec:pipeline:proposals} are stated as rules but not as numbers, so this step does not reconstruct the bands: the per-rule bonuses, the definition and trigger toll of the high-toll anchor, the guard demotion mechanics and the band cut-off are not published, and \texttt{occurrence\_merges.csv} in the snapshot is therefore the authority for the status and evidence string of every proposed pair rather than a check on a re-derivation.
\item \textbf{Canonical values.} Take the record from the highest-ranked source (Section~\ref{sec:pipeline:priority}: rank~3 uniform authorities, rank~2 MAK and BAAA, rank~1 Wikidata, ASN and SKYbrary), then fill its null fields from the records it beat. The rank is a property of the source code, not of the row, so it is not a column of \texttt{event\_sources.csv}: join on \texttt{source} to obtain it. Break ties by first arrival, except among the three tier-aware sources of Section~\ref{sec:pipeline:linkage}, where a final report outranks an interim one and an interim a preliminary; publication date plays no part.
\item \textbf{Derived variables.} \texttt{severity} $=$ \texttt{fatal} if \texttt{fatalities\_total} $> 0$, else per the rule in Section~\ref{sec:pipeline:mapping}; \texttt{region} is not derived from anything --- it is null on every row of the published corpus and no step in this protocol produces it.
\item \textbf{Table~\ref{tab:sources} and~\ref{tab:composition}.} For Table~\ref{tab:sources}, group \texttt{event\_sources.csv} by \texttt{source} and count rows, taking the year window from the joined occurrences. For Table~\ref{tab:composition}, group \texttt{events.jsonl} by decade of \texttt{event\_date} and by \texttt{country\_iso} collapsed to three columns --- \texttt{US}, any other code, and null --- keeping null as its own column rather than distributing it. Neither table is grouped by region; see Section~\ref{sec:limitations:fields}.
\item \textbf{Linkage evaluation.} Load \texttt{occurrence\_merges.csv} (snapshot), \val{4{,}068} rows with columns \texttt{(absorbed\_id,\allowbreak survivor\_id,\allowbreak status,\allowbreak method,\allowbreak score,\allowbreak evidence,\allowbreak decided\_by,\allowbreak decided\_at)}: \val{3{,}441} rows are \texttt{auto} and \val{627} \texttt{pending}. The \texttt{decided\_by} and \texttt{decided\_at} columns are empty on every row --- the pipeline never populated them, so the merge table carries no adjudication trail and the audit trail is \texttt{adjudicated\_pairs.csv} alone; all adjudication was LLM-assisted under the author's supervision, as declared under Assisted work. There is no weight file and no threshold to sweep, and the band assignment cannot be reproduced from this paper (step~7): take the \texttt{status} column of \texttt{occurrence\_merges.csv} as given and reproduce the reported error rates by comparing the five adjudicated samples against those bands as in Table~\ref{tab:linkage}. The file is consistent with a cut-off of \val{0.80} on \texttt{score} combined with the fan guard --- the lowest \texttt{auto} score is \val{0.80} and all \val{270} \texttt{pending} rows at or above it carry \texttt{fan-demoted} --- but that is an inference from the deposit, not a published rule. Sample~4 is additionally re-drawable: take the \val{3{,}441} rows with status \texttt{auto} in the order they appear in the file and draw \val{60} with Python's \texttt{random.sample} after \texttt{random.seed(\val{20260901})}; the draw is generator-specific, so another language's RNG seeded with the same number selects a different set. Those samples are deposited as \texttt{adjudicated\_pairs.csv}, \val{200} rows with columns \texttt{(sample,\allowbreak pair\_id,\allowbreak absorbed\_id,\allowbreak survivor\_id,\allowbreak event\_date,\allowbreak fatalities,\allowbreak verdict,\allowbreak basis)}; the error rates in Table~\ref{tab:linkage} are counts over that file and need no estimator. There is no cluster file and no stratified interval to compute --- see Section~\ref{sec:validation:linkage} for why neither exists. Figure~\ref{fig:weights} is not built from these files: it counts sources per occurrence in \texttt{event\_sources.csv.gz} and is described there because it stands in place of the match-weight histogram this design cannot produce.
\item \textbf{Coverage against ASN and B3A.} Not reproduced --- see Section~\ref{sec:validation:coverage} for why the comparison is not reported: both compilations are ingested by the corpus, so the check would measure a subset of the corpus against itself.
\item \textbf{Overlap matrix.} From \texttt{event\_sources.csv}, self-join on \texttt{occurrence\_id} to count occurrences per source pair; order by contribution and show raw counts, the diagonal being each source's total; Figure~\ref{fig:overlap}.
\item \textbf{Internal checks.} Run \texttt{node describe.mjs} (snapshot) to regenerate \texttt{descriptives.json}, which carries the completeness and consistency statistics of Section~\ref{sec:validation:internal}. It reads \texttt{events.jsonl.gz} and \texttt{event\_sources.csv.gz} and uses the Node standard library only; \texttt{make\_figures.mjs} regenerates the three figures from the same two files.
\end{enumerate}

\section{Datasheet summary}
\label{app:datasheet}

Following \citet{gebru2021datasheets}, this appendix is the dataset's datasheet; the deposit carries no separate datasheet file. \emph{Motivation:} to provide a provenance-preserving, cross-jurisdictional occurrence dataset for safety research; created by the author; no external funding. \emph{Composition:} \val{279{,}485} occurrence records and \val{131{,}882} attached narrative records (\val{164{,}068} in the narrative store overall); instances are investigated or reported aviation occurrences; the dataset is a sample of the population of occurrences, biased as described in Section~\ref{sec:data:biases}; narratives may contain names of individuals as published by the source authorities; no labels beyond the harmonized fields. \emph{Collection:} retrieved from public official sources by scripted adapters, most recently \val{2026-08}; no individuals were contacted. \emph{Preprocessing:} as in Section~\ref{sec:pipeline}; source records are held in a staging table at their state as of the last re-sync, and earlier states are not retained. \emph{Uses:} safety statistics, source-comparison studies, text mining; not suitable for rate estimation without external denominators or for identifying individuals. \emph{Distribution:} API and Zenodo snapshot; per-source licences. \emph{Maintenance:} re-ingested on per-source schedules; versioned releases with changelogs; errata accepted at the corresponding author.

\bibliographystyle{unsrtnat}
\bibliography{refs}
\end{document}